%% file: template.tex
\documentclass{article}

\usepackage{arxiv}

\usepackage[utf8]{inputenc} 
\usepackage[T1]{fontenc}    
\usepackage{hyperref}       
\usepackage{url}            
\usepackage{booktabs}       
\usepackage{amsfonts}       
\usepackage{nicefrac}       
\usepackage{microtype}      
\usepackage{lipsum}
\usepackage{graphicx}
\usepackage{amsmath}
\usepackage{subfigure}
\usepackage{multirow}
\usepackage{algorithm}
\usepackage{algorithmic}

\usepackage{xcolor}
\usepackage[normalem]{ulem}

\definecolor{forestgreen}{rgb}{0.13, 0.54, 0.13}

\title{Coupled and Uncoupled Dynamic Mode Decomposition in Multi-Compartmental Systems with Applications to Epidemiological and Additive Manufacturing Problems}

\author{
   Alex Viguerie \\
  Department of Mathematics\\
  Gran Sasso Science Institute\\
  Viale Francesco Crispi 7, L'Aquila, AQ 67100, Italy \\
  \texttt{alexander.viguerie@gssi.it} \\
  \And
    Gabriel F. Barros \\
  Dept. of Civil Engineering\\
  COPPE/Federal University of Rio de Janeiro \\
  P.O. Box 68506, RJ 21945-970, Rio de Janeiro, Brazil \\
  \texttt{gabriel.barros@coc.ufrj.br} \\
     \And
  Malú Grave \\
  Dept. of Civil Engineering\\
  COPPE/Federal University of Rio de Janeiro \\
  P.O. Box 68506, RJ 21945-970, Rio de Janeiro, Brazil \\
  \texttt{malugrave@nacad.ufrj.br} \\
   \And
 Alessandro Reali \\
  Dipartimento di Ingegneria Civile ed Architettura\\
  Università di Pavia \\
  Via Ferrata 3, Pavia, PV 27100, Italy \\
  \texttt{alereali@unipv.it} \\
   \And
 Alvaro L.G.A. Coutinho \\
  Dept. of Civil Engineering\\
  COPPE/Federal University of Rio de Janeiro \\
  P.O. Box 68506, RJ 21945-970, Rio de Janeiro, Brazil \\
  \texttt{alvaro@nacad.ufrj.br} \\
}

\begin{document}

\maketitle

\begin{abstract}
Dynamic Mode Decomposition (DMD) is an unsupervised machine learning method that has attracted considerable attention in recent years owing to its equation-free structure, ability to easily identify coherent spatio-temporal structures in data, and effectiveness in providing reasonably accurate predictions for certain problems, particularly over short-to-medium time frames. Despite these successes, the application of DMD to certain problems featuring highly nonlinear transient dynamics remains challenging. In such cases, DMD may not only fail to provide acceptable predictions but may indeed fail to recreate the data in which it was trained, restricting its application to diagnostic purposes (i.e., feature identification and extraction). For many such problems in the biological and physical sciences, the structure of the system obeys a compartmental framework, in which the transfer of mass within the system moves within states. In \textcolor{black}{these} cases, the behavior of the system may not be accurately recreated by applying DMD to a single quantity within the system, as proper knowledge of the system dynamics, even for a single compartment, requires that the behavior of other compartments \textcolor{black}{is} taken into account in the DMD process. In the present work, we demonstrate, theoretically and numerically, that\textcolor{black}{,} when performing DMD on a fully coupled PDE system with compartmental structure, one may recover useful predictive behavior, even when DMD performs poorly when acting compartment-wise. We also establish that important physical quantities, such as mass conservation, are maintained in the coupled-DMD extrapolation. The mathematical and numerical analysis suggests that DMD, properly applied, may be a powerful tool when applied to this common class of problems. \textcolor{black}{In particular, we show interesting numerical applications to a continuous delayed-SIRD model for Covid-19, and to a problem from additive manufacturing considering a nonlinear temperature field and the resulting change of material phase from powder, liquid, and solid states.}
\end{abstract}

\input{sections/1_introduction}
\input{sections/2_fem_dmd}

\input{sections/3_methodology}
\input{sections/4_numerical}
\input{sections/5_SEIRD}
\input{sections/6_AM}

\input{sections/8_conclusions}

\bibliographystyle{abbrv}  
\bibliography{references} 
\end{document}

%% file: sections/1_introduction.tex
\section{Introduction}
\par 
In recent years, data-driven techniques have become ubiquitous in science and engineering, as large quantities of high-dimensional data become more available and easier to handle on modern computer architectures \cite{Brunton2019book}. Though algorithms and techniques for the numerical simulation of physical systems have improved, such simulations still demand considerable resources in terms of both time and computing power. For such problems, the emerging field of Scientific Machine Learning (SciML) represents a potential way to reduce such costs by exploiting the inherent structure of such large datasets. Such techniques can be used in lieu of simulation or as a tool to extract useful information from numerical simulation data. SciML can be useful, for instance, to accelerate convergence, extract long-term behavior using only short-term simulations, or reduce the number of forward solves in applications involving the solution of hundreds, or possibly thousands, of numerical simulations (such as those common in optimization or uncertainty quantification). 
\par 
  There \textcolor{black}{is} a wide range of SciML methods, which have different theoretical bases and intended use cases \cite{Brunton2020}. \textcolor{black}{For more information on the field of SciML as a whole, we kindly refer the reader to \cite{Tajdari2021, Guo2019, Raissi2019, Brunton2020, Li2021, Li2020}.} In the present work, we discuss Dynamic Mode Decomposition (DMD), an unsupervised SciML method that exploits the underlying low-rank structure in many large datasets in order to extract only the most relevant information. This low-rank reconstruction may then be \textcolor{black}{employed for diagnostic purposes or}  to make future state predictions. While DMD may be used on both numerical \textcolor{black}{simulations} \cite{Schmid2010} and experimental data \cite{Schmid2011}, \textcolor{black}{in the present work}, we restrict our attention to the former, though the discussion applies equally well to experimental data. 
\par 
The standard DMD procedure takes $m$ snapshots of state measurements, \textcolor{black}{and uses them} to build a linear operator that best approximates the nonlinear system dynamics. While this idea has a strong theoretical motivation, in practice, DMD often fails to provide adequate results for extrapolation and prediction. This DMD behavior is particularly common for problems exhibiting \textcolor{black}{a} highly nonlinear transient behavior. In such cases, while DMD may still provide valuable diagnostic information \cite{Brunton2019book, bai2020dynamic, Proctor2015, kim2021discovering}, it may fail to provide \textcolor{black}{an} acceptable predictive behavior, in some cases failing even to provide an accurate reconstruction of the training period. Attempts to overcome these shortcomings have been addressed in, e.g., \cite{sashidhar2021bagging, dylewsky2020dynamic, lu2020prediction}, in which various different methods, including statistical procedures, time delay embeddings, and augmenting the observable space with additional nonlinear variables have been proposed to improve such performance.  
\par 
\textcolor{black}{Types} of system for which DMD may fail to provide acceptable predictions are \textcolor{black}{those of reaction-diffusion type, commonly found in biological and physical sciences} \cite{lu2020prediction, BGVRC2021, DMDSolids}. Such systems often feature a compartmental structure, in which the total mass of the system is decomposed into various compartments (or species) \cite{viguerie2020diffusion}. In the special case of closed compartments, the total mass of the system remains constant, with no additional sources or sinks. For large scale-systems, DMD may be performed compartment-wise, applying the technique on each physical compartment separately \cite{BGVRC2021}. Such an approach is understandable in the presence of many compartments and large-scale numerical discretizations, owing to the reduced costs of performing such discretizations separately, possibly in parallel. However, while this method may still provide acceptable results for diagnostics and feature identification, it can provide suboptimal results when used for prediction, or even simply reconstruction, of the desired system. In particular, many important system-wide \textcolor{black}{characteristics}, such as conservation of physical quantities across compartments, may no longer be respected \cite{BGVRC2021}.
\par 
In the present work, we \textcolor{black}{advocate a fully-coupled approach}; that is, DMD should be performed on the entire compartmental system, rather than \textcolor{black}{on the single compartments individuall}y. Applying the technique in this manner may provide an advantage in system reconstruction and prediction, as knowledge of even a single compartment is inherently linked to the evolution of the other compartments. When performing DMD in this manner, the underlying structure governing the entire system may be properly taken into account and reconstructed, even when extrapolating the system in time to predict future states. Notably, for problems featuring a closed-compartment structure, in which \textcolor{black}{a} total quantity of the system remains constant in time, this property is maintained in the DMD reconstruction. Thus, for such a class of problems, we show that DMD, when applied properly, is particularly well-suited to provide reasonable forecasts.

\par
This paper is structured as follows: Section \ref{sec:fem_dmd} describes the relation between the discretization of multi-compartment reaction-diffusion PDEs in space and time and the DMD algorithm. Section \ref{sec:methodology} describes the proposed methodology of performing DMD on the fully coupled system, rather than \textcolor{black}{on} the individual compartments, and provides some theoretical results justifying this choice. In Section \ref{sec:numerical} we describe the numerical applications in this study: the use of DMD on a continuous delayed-SIRD model from epidemiology, and a model from additive manufacturing which considers a nonlinear temperature field and the resulting change of material phase from powder, liquid, and solid states. We demonstrate how compartment-by-compartment reconstructions both fail to provide useful predictions and violate the conservation of mass. We then show that performing DMD on the fully coupled system eliminates these problems, allowing for the recovery of useful extrapolations while respecting the problem physics. In Section \ref{sec:conclusions}, we draw our final remarks and conclusions.


    
    

%% file: sections/2_fem_dmd.tex
\section{Partial Differential Equations and Dynamic Mode Decomposition}
\label{sec:fem_dmd}
\par 
The use of numerical methods \textcolor{black}{for} the approximation of partial differential equations (PDEs) is widely considered in many research topics and industrial applications. Many natural and artificial phenomena can be mathematically modeled using PDEs, and, by solving these equations, we can describe and analyze multiple processes that help \textcolor{black}{improving the} society as a whole, from finances and social behavior to complex physical and engineering systems. Many numerical methods such as \textcolor{black}{, e.g.,} the finite difference method, \textcolor{black}{the} finite element method, and \textcolor{black}{the} boundary element method are reliable mathematical tools for the approximation of PDEs, allowing \textcolor{black}{a} proper quantification of \textcolor{black}{the different} quantities of interest.
\par In the present study, we focus on using the finite element method for \textcolor{black}{the spatial discretization of} a system of compartmental reaction-diffusion equations. That is, consider a system of generic transient parametric PDEs for a vector $\mathbf{u}=\lbrack u_1,\,u_2,\,...,u_n\rbrack^T$, where the $u_i$ correspond to the vector compartments. We define \textcolor{black}{$\boldsymbol{\zeta}$} as a vector of parameters governing the control of the nonlinear system, and $\mathbf{f} = \lbrack f_1(t;\textcolor{black}{\boldsymbol{\zeta}}),\,f_2(t;\textcolor{black}{\boldsymbol{\zeta}}),...,\,f_n(t;\textcolor{black}{\boldsymbol{\zeta}}) \rbrack^T$ as given functions. We then consider the equation:  \begin{equation}\begin{split}
    \dfrac{\partial \mathbf{u}}{\partial t} + \mathcal{N}(\mathbf{u};\textcolor{black}{\boldsymbol{\zeta}}) =\mathbf{f}, \text{ \hspace{2cm} in $\Omega \times (0, T]$}, \\  
    \label{eq:strongVector}
\end{split}\end{equation}
\noindent with
 $$ \mathcal{N} = \begin{pmatrix} \mathcal{N}_{1,1} & \mathcal{N}_{1,2} & ... &\mathcal{N}_{1,n} \\
            \mathcal{N}_{2,1} & \mathcal{N}_{2,2}  & ... & \mathcal{N}_{2,n} \\
            \vdots & \vdots & \ddots & \vdots \\
            \mathcal{N}_{1,n}& \hdots & \hdots &  \mathcal{N}_{n,n} \\ \end{pmatrix},
$$
where the subscripts $i,j$ indicate that the operator $\mathcal{N}_{i,j}$ defines the interaction between the $i$-th and $j$-th \textcolor{black}{compartments}.
\noindent
\noindent The equation is equipped with boundary and initial conditions for each compartment
\begin{equation}
    \begin{aligned}
        \mathbf{u} &= \mathbf{g} \text{ \hspace{4cm} on $\Gamma_D \times (0, T]$,} \\
        \nabla \mathbf{u} \cdot \mathbf{n} &= \mathbf{h} \text{ \hspace{4cm} on $\Gamma_N \times (0, T]$,} \\
        \mathbf{u}(\mathbf{x},0;\textcolor{black}{\boldsymbol{\zeta}}) &= \mathbf{u}_0(\mathbf{x};\textcolor{black}{\boldsymbol{\zeta}}) \text{ \hspace{3.85cm} on $\tilde{\Omega}$},
    \end{aligned}
\end{equation}
where $\mathbf{g}=\lbrack g_1,\,g_2,\,...,g_n\rbrack^T$ and $\mathbf{h} =\lbrack h_1,\,h_2,\,...,h_n\rbrack^T$ are the vectors containing the Dirichlet and Neumann boundary conditions for each equation, respectively, $T$ is the final time, and $\mathbf{u}_0 = =\lbrack {u_0}_1,\,{u_0}_2,\,...,{u_0}_n\rbrack^T$ are the initial conditions for $\mathbf{u}$. The equations are solved on a bounded domain $\tilde{\Omega}$ composed by the domain $\Omega \subset \mathbb{R}^{nsd}$ and Lipschitz continuous boundaries $\Gamma_D \cup \Gamma_N = \Gamma \subset \mathbb{R}^{nsd - 1}$. Also, $\mathbf{n}$ is the unit outward normal to $\Gamma_N$. Considering the classical finite element method for the spatial discretization of the PDEs, the bounded domain is split into a finite number of elements with non-overlapping domain $\Omega^e \subset \mathbb{R}^{nsd}$ and boundaries $\Gamma^e \subset \mathbb{R}^{nsd-1}$. \textcolor{black}{Moreover}, the finite element method requires the equations to be converted in their weak forms by integrating Eq. (\ref{eq:strongVector}) against a weighting function $\mathbf{w} \in H^1(\Omega)$ and, then, applying the divergence theorem on the resulting equation. The weighting functions \textcolor{black}{exist} on $H^1(\Omega)$ which is the Sobolev space of the square-integrable functions with an integrable first weak derivative, allowing the weak form to be continuous. Considering $P^k(\Omega^e)$ the space of polynomials of degree equal or less than $k$ over $\Omega^e$, the function spaces are defined as: 
\begin{flalign}
    \label{eq:spaces}
    S_t^h &= \{\mathbf{u}^h(\cdot,t) \in H^1(\Omega) ~|~ \mathbf{u}^h(\cdot,t)|_{\Omega_e}\in P^k(\Omega^e), \forall e\}, \\
    W^h &= \{\mathbf{w}^h\in H^1(\Omega) ~|~ \mathbf{w}^h|_{\Omega_e}\in P^k(\Omega^e), \forall e\}.
\end{flalign}
That said, the weak formulation for the spatial discretization for the nonlinear system described on Eq. (\ref{eq:strongVector}) is: \textcolor{black}{Find} $\mathbf{u}^{h}(t) \in S_t^h$ such that  $\forall \mathbf{w}^h \in W^h$:
\begin{equation}
    \bigg(\dfrac{\partial \mathbf{u}^h}{\partial t}, \mathbf{w}^h\bigg) + (\mathcal{N}(\mathbf{u}^h; \mathbf{\zeta}), \mathbf{w}^h) = (\mathbf{f}^h, \mathbf{w}^h),
    \label{eq:weak}
\end{equation}
where the $\mathcal{L}^2$ inner product over the domain $\Omega$ is indicated by $(\cdot, \cdot)$. The choice for number of elements, element type, finite element formulation (i.e., Galerkin, stabilized formulations, etc.) is of major importance and depends largely on the nature of the problem to be solved. For instance, if the reaction term on the diffusion-reaction equations \textcolor{black}{is} dominant, the standard Galerkin formulation may fail to produce physical results, being well replaced by stabilized formulations such as the SUPG \cite{Brooks} 
or Variational Multiscale methods \cite{hughes,rasthofer,ahmed2017,codina2018,bazilevs2013computational}. For simplification regarding the notation, the superscript "$h$" will be omitted as the discretization of continuous variables has been applied. 

\par Since the system of equations has been partially approximated, temporal discretization of Eq. \eqref{eq:weak} is required for the complete resolution of the problem. From this point, the equation can be translated into a discrete-time dynamical system. In this system, the state vector $\mathbf{u}$ at the time instant $k+1$ can be written such that:
\begin{equation}
    \mathbf{u}_{k+1} = \mathcal{F}(\mathbf{u}_k)
    \label{eq:dynamical}
\end{equation}
that is, the system can now be considered as a temporal update of the state vector $\mathbf{u}_{k+1}$ containing the information of all the compartments on a given instant at a discrete level. The matrix $\mathcal{F}$ is the discrete-time flow map of the system and contains information regarding the parameters \textcolor{black}{$\boldsymbol{\zeta}$}, mesh size, solver properties, etc. The construction of $\mathbf{u}_{k+1}$ is done sequentially with the information of the previous state vector. Considering the evolution in time of a discretized PDE as a dynamical system is fundamental for the DMD application on numerical simulations.
\par 
DMD is a data-driven method able to extract the most dynamically relevant structures (DMD modes) on a spatio-temporal dataset, where each DMD mode is associated with frequency, and amplification/damping terms \cite{Kutz2016book}. The coherent structures are related in terms of space and time. The method is completely equation-free and data-driven, meaning that little to no assumptions on data must be considered for its applications. DMD was first applied in fluid dynamics applications \cite{Rowley2009,Schmid2010}, being expanded to many other applications such as epidemiology \cite{Proctor2015, BGVRC2021},  urban mobility \cite{Alla2020},  biomechanics \cite{Calmet2020}, climate \cite{Kutz2016} and aeroelasticity \cite{Fonzi2020}, especially in structure extraction from data and control-oriented methods. The method consists of creating a linear map of the dynamics of a given spatio-temporal dataset, even if the dynamics \textcolor{black}{is} nonlinear, by projecting the finite-dimensional nonlinear data using an infinite dimension operator able to linearly represent the flow map present on \eqref{eq:dynamical} for all time steps.

\par 
Looking at transient PDEs through the lens of dynamical systems is crucial for the application of DMD and the system described on Eq. (\ref{eq:dynamical}) can be considered for this purpose. \textcolor{black}{Let $\mathbf{Y}_j$ be a dataset} containing the state vectors for the compartment $j$ resulting of the dynamical system $\mathbf{u}_j$ for $k = 0, 1, \dotsc, m$, where $m+1$ is the total number of entries. The dataset
\begin{equation}
                \mathbf{Y}_j = \left[
                \begin{array}{cccc}
                \vrule & \vrule &        & \vrule \\
                \mathbf{u}^0_j  & \mathbf{u}^1_j  & \ldots & \mathbf{u}^m_j    \\
                \vrule & \vrule &        & \vrule
                \end{array}
                \right]
\end{equation}
can be rearranged into two matrices $\mathbf{Y}'_j = [\mathbf{u}^0_j \dotsc \mathbf{u}^{m-1}_j]  \in \mathbb{R}^{n \times m}$ and $\mathbf{Y}_j'' = [\mathbf{u}^1_j \dotsc \mathbf{u}^m_j]  \in \mathbb{R}^{n \times m}$, where $n$ is the number of nodes in the mesh.   
DMD aims \textcolor{black}{at} constructing the best fit approximation of the linear operator $\mathbf{A}$ that maps dataset $\mathbf{Y}'$ into dataset $\mathbf{Y}''$, that is, 
\begin{equation}
    \mathbf{Y}'' = \mathbf{A}\mathbf{Y}'.
\end{equation} 

The matrix $\mathbf{A}$ can be computed as $\mathbf{A} = \mathbf{Y}''\mathbf{Y}'^{\dagger}$, where $\mathbf{Y}'^{\dagger}$ is the Moore-Penrose pseudoinverse of $\mathbf{Y}'$. However, we avoid the computation of the full matrix $\mathbf{A}$ since \textcolor{black}{it} is a $n \times n$ matrix. Also, the computation of the full Moore-Penrose pseudoinverse is not advisable due to its ill-conditioning. Instead, we can compute the SVD of $\mathbf{Y}'$ as 
\begin{equation}
\label{eq:SVD}
    \mathbf{Y}'= \mathbf{U\Sigma V}^T,
\end{equation}
where  $\mathbf{U} \in \mathbb{R}^{n \times m}$ and $\mathbf{V} \in \mathbb{R}^{m \times m}$ are the left and right singular vectors and $\mathbf{\Sigma} \in \mathbb{R}^{m \times m}$ is a diagonal matrix with real, non-negative, and decreasing entries named singular values. The singular values $\sigma_0 \geq \sigma_1 \geq \sigma_2 \geq \dots \geq \sigma_{m-1}$ are hierarchical and can be interpreted in terms of how much the singular vectors influence the original matrix $\mathbf{Y}'$. For the DMD procedure, considering the Eckart-Young Theorem \cite{Eckart1936}, the optimal low-rank update approximation matrix $\mathbf{Y}'$, when subjected to a truncation rank $r$, can be written as    
\begin{equation}
    \mathbf{Y}' \approx \mathbf{\tilde{Y}}' = \mathbf{U}_r\mathbf{\Sigma}_r\mathbf{V}^T_r,
\end{equation}
where $\mathbf{U}_r \in \mathbb{R}^{n \times r}$ is a matrix containing the first $r$ columns of $\mathbf{U}$, $\mathbf{V}_r \in \mathbb{R}^{m \times r}$ contains the first $r$ columns of $\mathbf{V}$, and $\mathbf{\Sigma}_r \in \mathbb{R}^{r \times r}$ is the diagonal matrix containing the first $r$ singular values. The pseudoinverse can be approximated as
\begin{equation}
    \mathbf{Y}^{\dagger}{}' \approx \mathbf{\tilde{Y}}^{\dagger}{}'= \mathbf{V}_r\mathbf{\Sigma}_r^{-1}\mathbf{U}_r^T
\end{equation}
and, instead of computing $\mathbf{A}\in \mathbb{R}^{n \times n}$, we can obtain $\mathbf{\tilde{A}}$, a $r \times r$ projection of $\mathbf{A}$, as
\begin{equation}
    \mathbf{\tilde{A}} = \mathbf{U}_r^T\mathbf{A}\mathbf{U}_r =  \mathbf{U}_r^T\mathbf{Y}''\mathbf{V}_r\mathbf{\Sigma}^{-1}_r.
\end{equation}
\par 
Note that $\mathbf{\tilde{A}}$ is unitarily similar to $\mathbf{A}$. Further mathematical details regarding the optimization problem (the best-fitting matrix $\mathbf{\tilde{A}}$) and the influence of the Eckart-Young Theorem on constraints of the problem can be found in \cite{heas2020lowrank}. Now we can compute the eigendecomposition of $\mathbf{\tilde{A}}$:
\begin{equation}
    \mathbf{\tilde{A}W = W\Lambda},
\end{equation}
where $\mathbf{\Lambda}$ is a diagonal matrix containing the discrete eigenvalues $\lambda_j$ and the matrix $\mathbf{W}$ contains the eigenvectors \textcolor{black}{$\boldsymbol{\phi}_j$} of $\mathbf{\tilde{A}}$. The DMD basis can be written as:
\begin{center}
	\begin{equation}
	\mathbf{\Psi} = \mathbf{Y}''\mathbf{V}_r\mathbf{\Sigma}^{-1}_r\mathbf{W},
	\end{equation}
\end{center}
and the discrete signal reconstruction as:

\begin{equation}\label{DMDExpansionDiscrete}
    \mathbf{u}_{k+1}  \approx  \tilde{\mathbf{u}}_{k+1} =  \mathbf{\Psi} \mathbf{\Lambda} \mathbf{b}.
\end{equation}
The continuous counterpart to \eqref{DMDExpansionDiscrete} reads:
\begin{equation}\label{DMDExpansionContinuous}
    \mathbf{u}(t)  \approx  \tilde{\mathbf{u}}(t) =  \mathbf{\Psi}\exp(\mathbf{\Omega}_{eig}t) \mathbf{b},
\end{equation}

with $\mathbf{b}$ being the vector containing the projected initial conditions such that $\mathbf{b} = \mathbf{\Psi^{\dagger}}\mathbf{u}_0$, and $\mathbf{\Omega}_{eig}$ is a diagonal matrix whose entries are the continuous
eigenvalues $\omega_i = \ln(\lambda_i)/\Delta t_o$, where $\Delta t_o$ is the time step size between the outputs. The form of \eqref{DMDExpansionContinuous} can be regarded as a generalization of the Sturm-Liouville expansion for a differential problem:
\begin{align}\label{SturmLiouville}
    u(t) &= \sum_{i=0}^{\infty} b_i \psi_i e^{\omega_i t},
\end{align}
where $\psi_i$ and $\omega_i$ are the $i$-th Sturm-Liouville eigenfunctions and eigenvalues for a given differential operator.

%% file: sections/3_methodology.tex
\section{DMD applied to Multi-Compartment Systems}
\label{sec:methodology}
Regarding the use of DMD on systems of the type \eqref{eq:strongVector}, one may employ one of two approaches. The first approach, which we will refer to \textcolor{black}{as} the \textit{uncoupled approach}, involves treating each $u_i$ separately in the DMD process. That is, we define the snapshot matrices:
\begin{align}\begin{split}\label{uncoupledDMD}
    \mathbf{Y}_1 &= \left[
                \begin{array}{cccc}
                \vrule & \vrule &        & \vrule \\
                u_1^0  & u_1^1  & \ldots & u_1^m   \\
                \vrule & \vrule &        & \vrule
                \end{array}
                \right] \\
                    \mathbf{Y}_2 &= \left[
                \begin{array}{cccc}
                \vrule & \vrule &        & \vrule \\
                u_2^0  & u_2^1  & \ldots & u_2^m   \\
                \vrule & \vrule &        & \vrule
                \end{array}
                \right] \\
                &\vdots \\
                    \mathbf{Y}_n &= \left[
                \begin{array}{cccc}
                \vrule & \vrule &        & \vrule \\
                u_n^0  & u_n^1  & \ldots & u_n^m   \\
                \vrule & \vrule &        & \vrule
                \end{array}
                \right].
\end{split}\end{align}
We then perform the DMD algorithm $n$ times, obtaining $n$ separate reconstructions, one for each compartment. This approach was used in \cite{BGVRC2021}, where it was shown that it does not always yield acceptable results\textcolor{black}{. In} particular, \textcolor{black}{in some situations,} we may observe large spurious oscillations and rapid deterioration of error  \cite{BGVRC2021}\textcolor{black}{, and, in extreme cases, it} may even fail to reproduce the dataset used for the training, as has been observed in many DMD applications \cite{sashidhar2021bagging}.
\par In contrast, we may consider the following alternative:
\begin{equation}\label{coupledDMD}
                \mathbf{Y} = \left[
                \begin{array}{cccc}
                u_1^0  & u_1^1  & \ldots & u_1^m    \\
                \vrule & \vrule &        & \vrule \\ 
                               u_2^0  & u_2^1  & \ldots & u_2^m    \\
                \vdots & \vdots &  \ddots      & \vdots \\
                                u_n^0  & u_n^1  & \ldots & u_n^m    \\
                \end{array}
                \right] = \left[
                \begin{array}{cccc}
                \vrule & \vrule &        & \vrule \\
                \mathbf{u}^0  & \mathbf{u}^1  & \ldots & \mathbf{u}^m    \\
                \vrule & \vrule &        & \vrule
                \end{array}
                \right].
\end{equation}
The DMD reconstruction is then performed once, on a larger system. To recover the signal reconstruction of the $i$-th compartment for a system containing $n$ compartments of dimension $s$, we recall that the DMD reconstruction has the same topology as the input data, and hence:
\begin{equation}\label{DMDExpansion}
    u_i (t)  \approx  \tilde{\mathbf{u}}\mathtt{\big\lbrack(i-1)s:is\big\rbrack} (t) =  \mathbf{\Psi}\exp(\mathbf{\Omega}_{eig}t) \mathbf{b}\mathtt{\big\lbrack(i-1)s:is\big\rbrack}.
\end{equation}

\par However, it is also the case that, even for large-scale PDE problems, the compartments $u_i$ are not, in fact, independent quantities; rather, they may interact with one another in a complex, nonlinear fashion. With this in mind, it makes sense that a proper reconstruction of a given compartment may require adequate knowledge of all of the other compartments in a nonlinear system.

\par In terms of computational aspects, the uncoupled or coupled approach should be considered carefully. While the overall amount of data to be processed is identical for both methods, computational and memory costs are not. For instance, loading the complete dataset when applying the coupled DMD for a sufficiently large number of compartments could be prohibitive on a given system, depending on the number of degrees of freedom and available memory. However, if this is the case, the uncoupled approach may alleviate these restrictions by loading one (or a batch of) \textcolor{black}{compartment} at a time from disk. As for computational time, the algorithmic complexity of DMD is primarily driven by the SVD algorithm. This dependence is inherited when evaluating the use of the uncoupled or coupled cases. The classical SVD algorithm \cite{heath_book} computes the singular values of a dense matrix at the cost of $\mathcal{O}(n^3)$, implying that the coupled approach will scale cubically with the number of compartments, while the uncoupled approach will scale linearly. This bad scaling leads to a prohibitively high cost of the coupled approach for a large $n$ and/or a large number of compartments.  

However, it is worth noting that the classical SVD algorithm considers the complete factorization of the dataset, while DMD only relies on the first $k$ singular values and vectors. In a situation with \textcolor{black}{a} large $n$ or a large number of compartments, the extra cost of the coupled DMD may be circumvented by the Randomized Singular Value Decomposition (rSVD) \cite{Halko2011, Erichson2019_rsvd}, which computes the approximate rank-$k$ SVD with a computational complexity of $\mathcal{O}(nmk)$. Applying the rSVD may enable coupled/uncoupled approaches to achieve a similar overall complexity. We note that there are many additional computational aspects to consider, but such an in-depth discussion is beyond the scope of the present work.  We also note that the uncoupled algorithm lends itself more naturally to parallelization.  We do acknowledge, however, that such aspects are important and certainly worthy of future investigation.   



\subsection{Mathematical analysis}
We now present three mathematical results suggesting that the coupled approach may provide more accurate and physically consistent reconstructions than the uncoupled \textcolor{black}{one} for closed-compartment systems. The first two theorems can be expected to hold for all such systems studied in the present work. The third requires more restrictive conditions which, while not guaranteed to be satisfied, we have found in practice \textcolor{black}{to be} often satisfied enough for the result to remain useful.
\newline \textbf{Theorem 1: } \textit{For a compartmental dynamical system $\mathbf{u} = \lbrack u_1,\,u_2,\,...,\, u_n \rbrack^T$ such that:}
 \begin{align}\label{condition1}  \int_{\Omega} \left( u_1 + u_2 + ... + u_n \right) d\Omega  &= M \qquad \forall t \in \{  t\, |\, t= j \Delta t\}_{j=0}^{\infty} .    \end{align}
\textit{the exact coupled DMD reconstruction $\widetilde{\mathbf{u}} = \lbrack \widetilde{u}_1,\,\widetilde{u}_2,\,...,\, \widetilde{u}_n \rbrack^T$ is also such that: }
\begin{align}\label{condition2}  \int_{\Omega} \left( \widetilde{u}_1 + \widetilde{u}_2 + ... + \widetilde{u}_n \right) d\Omega  &= M  \qquad \forall t \in \{ t\, |\, t= j \Delta t\}_{j=0}^{\infty}    \end{align}

\textbf{Proof. } 
From \eqref{condition1}, we may conclude that, for each instance $k$, and denoting the column vector of all ones as $\mathbf{1}$,
\begin{align}\label{constantSum} \mathbf{1}^{T}\mathbf{u}_ k &= C,    \end{align}
for some constant $C$\footnote{At the discrete level, in practice, this condition may require a certain degree of mesh-size uniformity.}. This implies that, for \eqref{coupledDMD}, 
\begin{align}\label{constantSumOnDMD}
    \mathbf{1}^T \mathbf{Y} &= \lbrack C,\,C,\,...,\,C\rbrack, 
\end{align}
and in particular, 
\begin{align}\begin{split}\label{ConstantSumY1Y2}
    \mathbf{1}^T \mathbf{Y}_1 &= \lbrack C,\,C,\,...,\,C\rbrack, \\
    \mathbf{1}^T \mathbf{Y}_2 &= \lbrack C,\,C,\,...,\,C\rbrack. \\
\end{split}\end{align}
We then have that:
\begin{align}\begin{split}\label{proofCore}
\mathbf{1}^{T}\widetilde{\mathbf{u}}_1 &= \mathbf{1}^T\mathbf{\Psi} \mathbf{\Lambda} \mathbf{b} \\
&= \mathbf{1}^T\mathbf{\Psi} \mathbf{\Lambda} \mathbf{\Psi}^{-1}\mathbf{u}_0 \\
&= \mathbf{1}^T \mathbf{A} \mathbf{u}_0 \\
&= \mathbf{1}^T \mathbf{Y}_2 \left(\mathbf{Y}_1\right)^{\dag}\mathbf{u}_0 \\
&= \lbrack C,\,C,\,...,\, C\rbrack^{T}\left(\mathbf{Y}_1\right)^{\dag}\mathbf{u}_0 \\
&= \mathbf{1}^T \mathbf{Y}_1 \left(\mathbf{Y}_1\right)^{\dag}\mathbf{u}_0 \\ 
&= \mathbf{1}^T \mathbf{u}_0  \\
&= C.
\end{split}\end{align}
This establishes the base case, for which one may apply mathematical induction for $\mathbf{u}_k$ using the exact argument above. We note that\textcolor{black}{,} in practice, we do not expect this condition to hold exactly, as we use approximations to $\mathbf{A}$. However, we should observe this behavior, at least approximately, in the coupled reconstruction. Although we obtained the result for the discrete time-flow map, we also expect it to hold for general $t$. A proof of this is desired.

 \noindent \textbf{Theorem 2. } \textit{For problems in which some subset of compartments satisfies the hypotheses of Theorem 1, Theorem 1 holds on the DMD reconstruction of the subset.} 
\newline \textbf{Proof. } As the ordering of compartments is arbitrary, reorder $\mathbf{Y}$ such that compartments $1,,\, 2,\,...,\, j$ are the first $j$ compartments. Then repeating the arguments of Theorem 1 with
$$ \widetilde{\mathbf{1}} = \begin{pmatrix} \mathbf{1}_1 \\ \mathbf{1}_2 \\ \vdots \\ \mathbf{1}_j \\ \mathbf{0}_{j+1} \\ \vdots \\ \mathbf{0}_n \end{pmatrix}  $$
instead of $\mathbf{1}$\textcolor{black}{,} establishes that the closed-compartment property holds on the subset. 
\par It remains to be shown that the compartment reordering only results in a similar reordering on the DMD reconstruction, and does not affect the reconstruction dynamics. Let $\mathbf{Y}$ denote the snapshot matrix in the original ordering of the system and $\widehat{\mathbf{Y}}$ the reordered snapshot matrix. Note that:
$$ \widehat{\mathbf{Y}} =  \mathbf{P} \mathbf{Y}, $$
where $\mathbf{P}$ is a permutation matrix. Then, 
\begin{align}\begin{split}\label{permutation}
    \widehat{\mathbf{Y}}_2 &=\widehat{\mathbf{A}}\widehat{\mathbf{Y}}_1 \\
    \mathbf{P}\mathbf{Y}_2 &= \widehat{\mathbf{A}}  \mathbf{P}\mathbf{Y}_1 \\
     \mathbf{P}\mathbf{Y}_2 \left(\mathbf{Y}_1\right)^{\dag}   \mathbf{P}^{-1} &= \widehat{\mathbf{A}} \\
     \mathbf{P}\mathbf{A}  \mathbf{P}^{-1} &= \widehat{\mathbf{A}}. \\
\end{split}\end{align}
Hence, $\mathbf{A}$ and $\widetilde{\mathbf{A}}$ are unitarily similar, and have the same eigenvalues. Further, for a given eigenvector $\mathbf{w}$ of $\mathbf{A}$, $\mathbf{P}\mathbf{w}$
is an eigenvector of $\widetilde{\mathbf{A}}$. The DMD modes are therefore identical up to the reordering permutation, yielding the desired result.

\textbf{Theorem 3. }  \textit{ For a compartmental problem which satisfies the hypotheses of Theorem 1, if we have additionally that $u_i \geq 0$, $\widetilde{u}_i \geq 0$ at all $(x,t)$ for each $i$, then the $L^1$ error of the DMD reconstruction: $$ \| \mathbf{u}- \widetilde{\mathbf{u}} \|_{L^1} $$ is bounded for all $t \in \{  t\, |\, t= j \Delta t\}_{j=0}^{\infty} $. }
\newline \textbf{Proof. } From the assumption of non-negativity on the compartments, it follows that:
\begin{equation}\begin{split}\label{positivity}
    \int_{\Omega}(u_1 + u_2 +... + u_n) d\Omega =      \int_{\Omega}| u_1 + u_2 +... + u_n | d\Omega = \| \mathbf{u} \|_{L^1}, \\
    \int_{\Omega}(\widetilde{u}_1 + \widetilde{u}_2 +... + \widetilde{u}_n) d\Omega =      \int_{\Omega}| \widetilde{u}_1 + \widetilde{u}_2 +... + \widetilde{u}_n | d\Omega = \| \widetilde{\mathbf{u}} \|_{L^1}. 
\end{split} \end{equation}
We then observe that:
\begin{align}\begin{split}\label{positivityThm}
    \| \mathbf{u} - \widetilde{\mathbf{u}} \|_{L^1} &\leq \| \mathbf{u} \|_{L^1} + \| \widetilde{\mathbf{u}} \|_{L^1}  \\
    &= 2M.
\end{split}\end{align}

\textbf{Remark 1. } Systems which satisfy the hypotheses of Theorems 1 and \textcolor{black}{2} are quite common, making these theorems valuable tools. Unfortunately, the non-negativity hypothesis on the $\widetilde{u}_i$ for Theorem 3 does not hold in general, and these quantities may become negative. In practice, we have found that this assumption remains satisfied for many problems, at least for a moderate period of time after the end of the training period. In such a regime, we may expect the error bound given by Theorem 3 to hold, preventing error blowup. These results suggest, however, that a variant of DMD guaranteeing non-negativity of $\widetilde{u}_i$ could be a significant development and is a direction worthy of investigation, as it would establish DMD as a valuable tool for predicting the dynamics of closed compartmental systems.

\textbf{Remark 2. } It is common in DMD to augment the variable space with nonlinear observables, typically functions of the state variables, to better capture the nonlinear dynamics of the system. Such \textcolor{black}{an} augmentation is motivated by the connection between DMD and the Koopman operator. We note that the result of Theorem 2 holds equally well if the variables which do not obey the closed-compartment structure are not necessarily compartments themselves but additional nonlinear variables. More concretely, one may consider a two-compartment model in which $$\int_{\Omega} (u_1 + u_2) = M$$ at each time-step. If one wishes to perform DMD on an augmented variable space $\lbrack u_1 ,\, u_2,\, g(u_1,u_2) \rbrack$, with $g$ some nonlinear function depending on $u_1$ and $u_2$, we would still expect that $$\int_{\Omega} (\widetilde{u}_1 + \widetilde{u}_2) = M,$$
despite the presence of the nonlinear variable.

%% file: sections/4_numerical.tex
\section{Numerical Experiments}
\label{sec:numerical}
\par In this section, we seek to understand the differences in accuracy between the coupled and uncoupled DMD formulations for both reconstruction as well as extrapolation and prediction. We select two different compartmental models in order to assess the effectiveness of the two approaches:

\begin{enumerate}
    \item A delayed-PDE SIRD model for COVID-19, which separates the total population into compartments based on their infection and immunity status;
    \item A problem inspired by additive manufacturing, consisting of a nonlinear thermal PDE coupled with a three-ODE compartmental model consisting of three differential equations in space, corresponding to the powder, liquid, and solid phases.
\end{enumerate}

We have selected these examples due to their \textcolor{black}{different} physics, application areas, and dynamics. They share, however, a mass-conservative closed-compartmental structure. Further details, including the formal mathematical definitions of each model, are given in the following.


%% file: sections/5_SEIRD.tex
\subsection{Delayed-SIRD Model}

The outbreak of COVID-19 in 2020 has led to a surge in interest in the mathematical modeling of infectious diseases. Disease transmission may be modeled  \textcolor{black}{by} \textit{compartmental models}, in which the population under study is divided into compartments and has assumptions about the nature and time rate of transfer from one compartment to another \cite{brauer2019mathematical}, owing to the original SIR framework of Kermack and McKendrick \cite{kermack1927contribution}\footnote{The standard SIR system is, in fact, a special case of a much more general renewal model introduced by Kermack and McKendrick. We refer the interested reader to \cite{breda2012formulation} for a more modern elucidation of this topic.}. While less common than the ordinary differential equation (ODE) version of such models, compartmental models for epidemic spread spatial transmission through the use of PDE models have also been applied to study both the COVID-19 epidemic as well as other diseases \cite{viguerie2020diffusion, Viguerie2021, guglielmi2021delay, GVBRC2021, bertrand2021least, bertaglia2021hyperbolic, boscheri2021modeling, jha2020bayesian,keller2013numerical, MurrayII}. 

Here, we work with a spatio-temporal delayed SIRD model, related to that presented in \cite{guglielmi2021delay} and given by the following set of equations:

\begin{equation}\begin{split}
&\frac{\partial s}{\partial t} + \beta_i \frac{si(t-\sigma)}{n_{pop}} + \beta_e \frac{si}{n_{pop}} - \nabla \cdot (n_{pop}\nu_s\nabla s) = 0
\end{split}
\label{covid_s_dens}
\end{equation}
\begin{equation}
\frac{\partial i}{\partial t} -  \beta_i \frac{si(t-\sigma)}{n_{pop}} - \beta_e \frac{si}{n_{pop}} + (\gamma  + \delta) i(t-\sigma) - \nabla \cdot (n_{pop}\nu_i \nabla i) = 0
\label{covid_i_dens}
\end{equation}
\begin{equation}
\frac{\partial r}{\partial t}  - \gamma i(t-\sigma)  -\nabla \cdot (n_{pop} \nu_r \nabla r) = 0
\label{covid_r_dens}
\end{equation}
\begin{equation}
\frac{\partial d}{\partial t} - \delta i(t-\sigma) = 0.
\label{covid_d_dens}
\end{equation}

\noindent where  \textcolor{black}{$s(\mathbf{x}, t)$, $i(\mathbf{x}, t)$, $r(\mathbf{x}, t)$, and $d(\mathbf{x}, t)$} denote the densities of the \textit{susceptible}, \textit{infected}, \textit{recovered}, and \textit{deceased} populations, respectively. The sum of all  \textcolor{black}{compartments} with the exception of $d(\mathbf{x},t)$ is represented by $n_{pop}$ which is the total living population. 
$\beta_i$ and $\beta_e$ denote the transmission rates between symptomatic and susceptible individuals and presymptomatic and susceptible individuals, respectively (units days$^{-1}$), $\gamma$ the symptomatic recovery rate (units days$^{-1}$), $\delta$ represents the mortality rate (units days$^{-1}$), and  \textcolor{black}{$\boldsymbol{\nu_s}$, $\boldsymbol{\nu_i}$, $\boldsymbol{\nu_r}$} are the diffusion \textcolor{black}{parameter matrices} of the different population groups as denoted by the sub-scripted letters (units km$^2$  persons$^{-1}$  days$^{-1}$). Note that all these parameters can be considered time and space-dependent. The time-delay $\sigma$ incorporates the lag effect present between the time of infection and development of symptoms, and seeks to model the delay observed in practice when new restrictions or other measures are introduced.  

Letting $\mathbf{u} = \lbrack s,\,i,\,r,\,d\rbrack^T $ and $\mathbf{\zeta}=\lbrack \beta_i,\,\beta_e,\,\gamma,\,\delta,\mathbf{\nu} \rbrack^T$ if,

\begin{equation}
   \mathbf{A} = \begin{bmatrix}
0 & 0 & 0 & 0 \\
0 & \gamma+\delta & 0 & 0 \\
0 & -\gamma & 0 & 0 \\
0 & -\delta & 0 & 0 \\
\end{bmatrix}
\end{equation}
\begin{equation}
   \mathbf{B} = \begin{bmatrix}
\frac{\beta_e}{n}i + \frac{\beta_e}{n}i(t-\sigma) & 0 & 0  &0 \\
-\frac{\beta_e}{n}i - \frac{\beta_e}{n}i(t-\sigma) & 0 & 0  & 0 \\
0 & 0 & 0 & 0 \\
0 & 0 & 0 & 0 \\
\end{bmatrix}
\end{equation}
\begin{equation}
   \boldsymbol{\nu} = \begin{bmatrix}
\boldsymbol{\nu_s} & 0 & 0 & 0  \\
0  & \boldsymbol{\nu_i} & 0 & 0\\
0  & 0 & \boldsymbol{\nu_r} & 0\\
0  & 0 & 0 & 0\\
\end{bmatrix}
\end{equation}
\begin{equation}
   \boldsymbol{\nu_k} = \begin{bmatrix}
 \nu^k_{xx} & \nu^k_{xy}\\
 \nu^k_{yx} & \nu^k_{yy} 
\end{bmatrix}
\textnormal{ with } k=s,i,r
\end{equation}
\begin{equation}
\mathcal{N}(\mathbf{u},\mathbf{\zeta}) = \mathbf{A}\mathbf{u}(t-\sigma) + \mathbf{B} \mathbf{u} - \nabla \cdot \left(n_{pop} \boldsymbol{\nu} \nabla \mathbf{u}\right)
\end{equation}

\begin{equation}
   \mathbf{f} = \begin{bmatrix}
0\\
0 \\
0 \\
0 \\
\end{bmatrix},
\end{equation}
we may write the system in the generic form given by \eqref{eq:strongVector}.
For the numerical solution of (\ref{covid_s_dens})-(\ref{covid_d_dens}), we discretize in space using a Galerkin finite element variational formulation. The resulting systems of equations are stiff, leading us to employ implicit methods for time integration. We apply the Backward Euler formula (BE), which offers first-order accuracy while remaining unconditionally stable. This choice is motivated by the ease of incorporating this time-stepping scheme with the presence of the delay term, which \textcolor{black}{is} known to make the use of multi-step methods more complex \cite{BZ03}. We implement the whole model in \texttt{libMesh} \cite{libmesh}. One may find more details about the methods in \cite{Viguerie2021, grave2020adaptive}.

\subsubsection{Simulation for Georgia}
\begin{figure}[ht!]
  \centering
 \includegraphics[width=\linewidth]{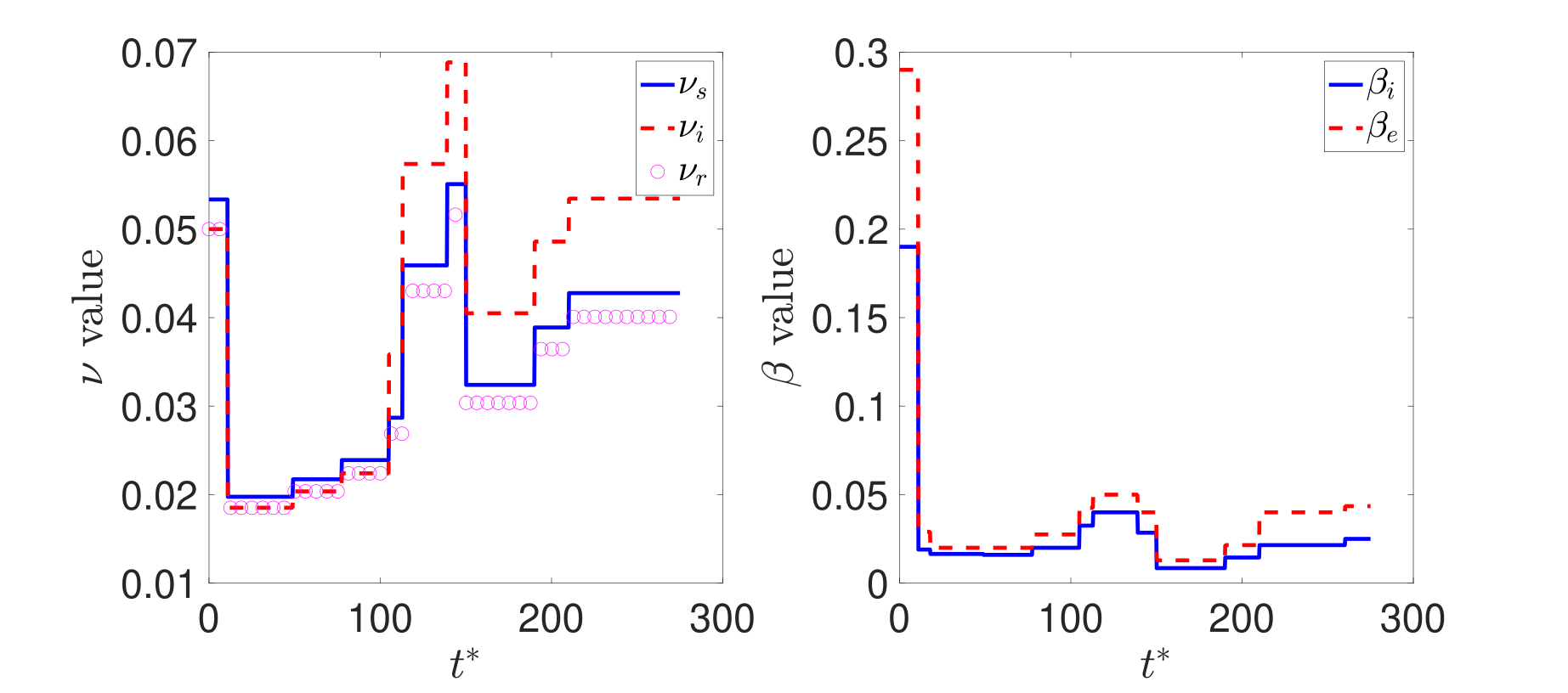}
  \caption{The changing values of $\nu$ and $\beta$ in time for the delayed-SIRD PDE simulation of Georgia. $\nu$ corresponds to mobility and $\beta$ to contact; the values were tuned empirically according to the imposition and relaxation of restrictions introduced by the state government, as well as mobility data and other significant events. }
  \label{fig:NuBeta}
\end{figure}

We now use the model \eqref{covid_s_dens}-\eqref{covid_d_dens} to simulate the outbreak of COVID-19 in the U.S. state of Georgia. The variables $\boldsymbol{\nu}$ and $\beta_{i,e}$ vary in time, and are shown in Fig. \ref{fig:NuBeta} \cite{GVBRC2021}. The other parameters \textcolor{black}{are assumed} fixed for simplicity, although an even more realistic model may incorporate time-dependence for these parameters as well.
\par We let $\delta=$ 1/180 days$^{-1}$, $\gamma$=1/24 days$^{-1}$ and $\sigma=7$ days. It was shown in \cite{guglielmi2021delay} that under the condition:
\begin{align}\label{bound1} \gamma + \delta < \frac{\pi}{2\sigma} \end{align}
we expect asymptotically stable behavior from the model \eqref{covid_s_dens}-\eqref{covid_d_dens}, and further under the more restrictive condition 
\begin{align}\label{bound2} \gamma + \delta < \frac{1}{e \sigma}, \end{align}
the compartments of the equations \eqref{covid_s_dens}-\eqref{covid_d_dens} should remain positive, preventing nonphysical oscillations into negative values. A straightforward calculation shows that\textcolor{black}{, for the adopted values of $\gamma$, $\delta$, and $\sigma$,} we may expect both stable and physical behavior.
\par The physical domain, mesh, and population distributions were reconstructed following the procedure outlined in \cite{GVBRC2021}. As this is a delay-differential equation model, we let $i(t)=i_0$ for $t \in \lbrack -\sigma,\,0 \rbrack$. We use a fixed time-step of $\Delta t=0.25$ days. Since $\sigma=7$ days is a fixed, exact multiple of the time-step size, we may directly use previously-computed solutions of $i$ for $i(t-\sigma)$, and do not need to use more sophisticated interpolation techniques. We note that extensions of \eqref{covid_s_dens}-\eqref{covid_d_dens} incorporating adaptive \textcolor{black}{time-stepping and/or a time- and state-dependent $\sigma$} will require a more involved computational procedure for $i(t-\sigma)$. We considered an unstructured triangular mesh consisting of 22310 elements, and used a piecewise linear finite element approximation.

\begin{figure}[ht!]
  \centering
 \includegraphics[width=.4\linewidth]{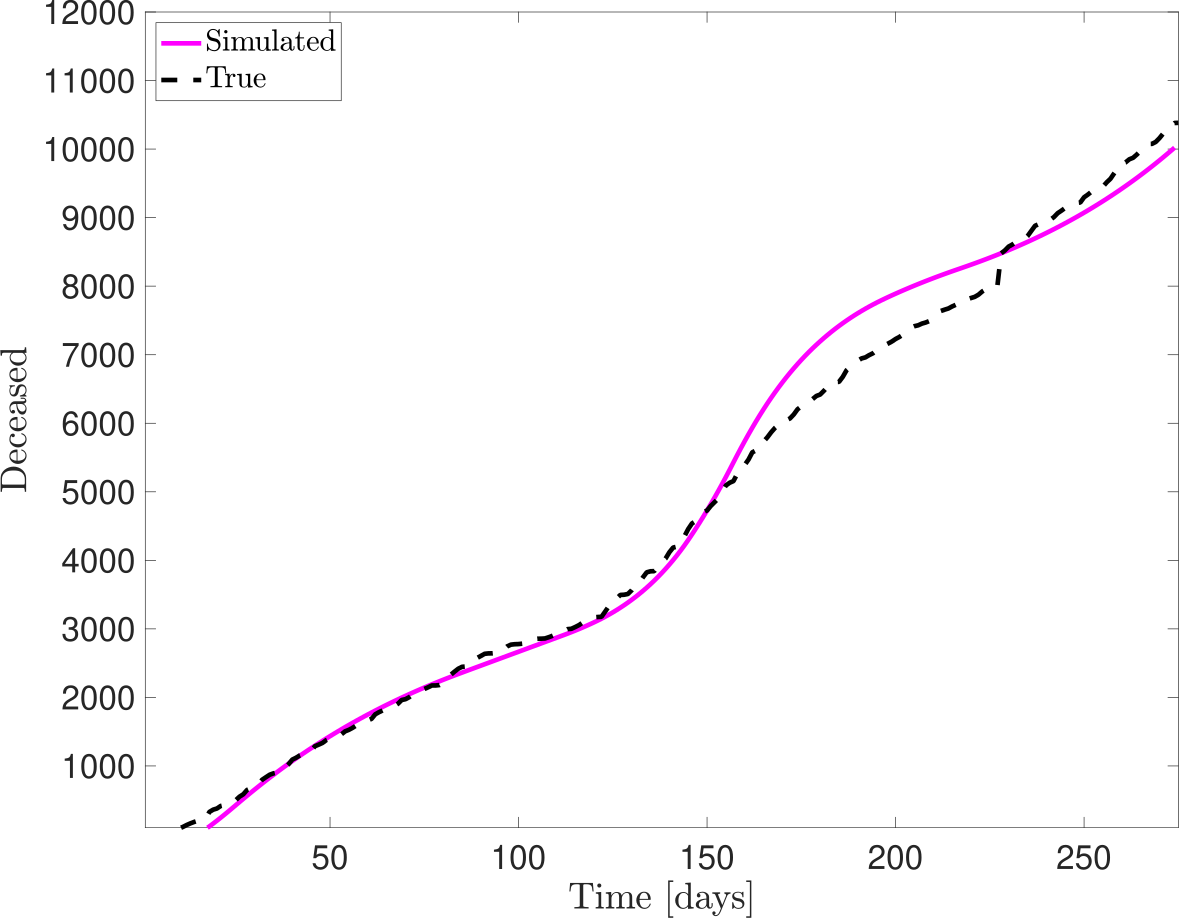}
  \caption{Fatalities in the state of Georgia; simulation versus measured data. We see very strong agreement between measurement and simulation.}
  \label{fig:ComparisonWMeasurement}
\end{figure}

\begin{figure}[ht!]
  \centering
 \includegraphics[width=\linewidth]{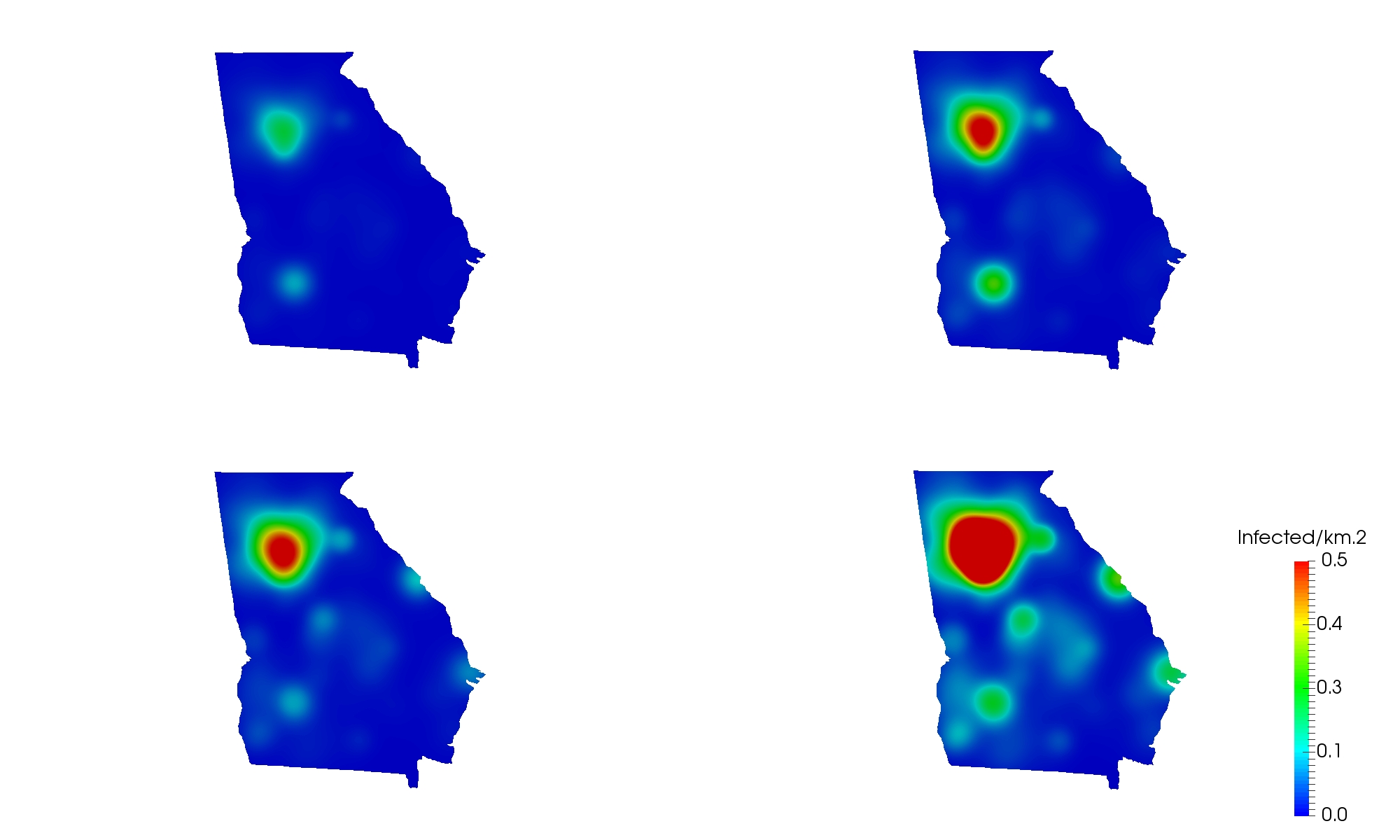}
  \caption{Evolution of the infected compartment $i$ for the Georgia simulation over different phases of the epidemic, from top-left, counter-clockwise: Day 1, 30, 120, 150. The outbreak severity is most pronounced in the Atlanta metropolitan areas, with other hotspots in Savannah (southeast), Augusta (east) and Albany (southwest) regions of the state. }
  \label{fig:GeorgiaDaybyDay}
\end{figure}
\par In Fig. \ref{fig:ComparisonWMeasurement}, we plot the results of the measured fatalities in Georgia against the simulated data. We observe excellent agreement overall, with an $R^2$ of 99.52\% \textcolor{black}{amd} a relative $L^2$ norm error of 5.74\%. We note that there is a jump in the measured data; this is due to previous unaccounted-for fatalities being retroactively added to the dataset. The evolution of the infected compartment $i$ over several different phases of the epidemic is shown in Fig. \ref{fig:GeorgiaDaybyDay}. 

\subsubsection{DMD Georgia Solution}
Having validated the baseline simulation at a satisfactory level, we now analyze the DMD reconstructions. We perform both the uncoupled and coupled DMD. In both cases, we train the DMD reconstruction using the first 250 days of the simulation, and then extrapolate over the following 25 days. Thus, the window for extrapolation is 10\% of the training period, consistent with the analysis performed in \cite{BGVRC2021}.
\par
\begin{figure}[ht!]
  \centering
 \includegraphics[width=.6\linewidth]{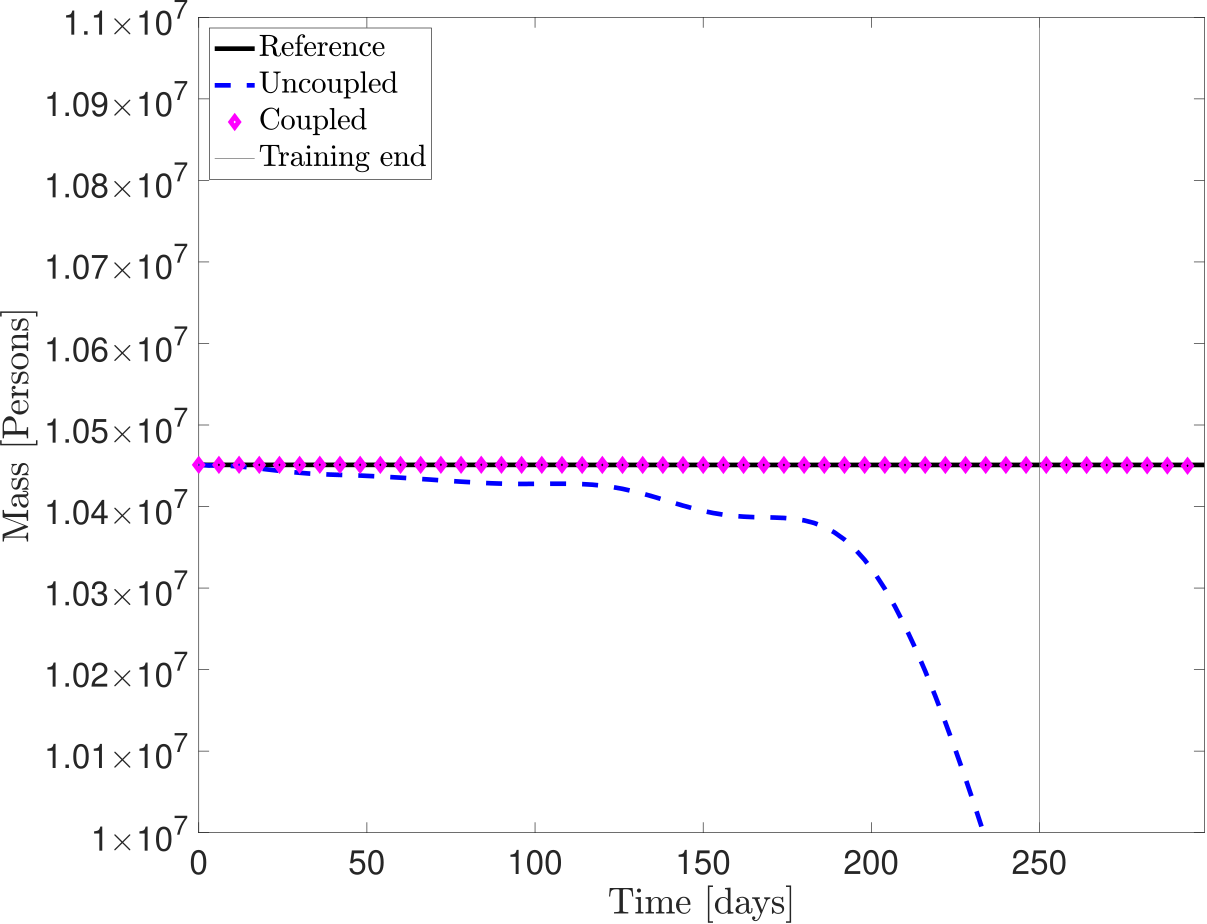}
  \caption{Comparison in conservation of mass between the coupled and uncoupled DMD reconstructions. We observe that the coupled reconstruction successfully conserves mass, even past the training period, while the uncoupled reconstruction does not respect this important property.}
  \label{fig:MassConsv}
\end{figure}

 We first confirm Theorem 1 numerically. As we do not consider new births or non-COVID deaths in \eqref{covid_s_dens}-\eqref{covid_d_dens}, we therefore expect \textcolor{black}{that}:
\begin{align}
    \frac{d}{dt}\left\lbrack \int_{\Omega} \left(s(t)+i(t)+r(t)+d(t) \right) d\Omega \right\rbrack &= 0 
\end{align}
for all $t$, representing a conservation of total mass in the system. We compute the quantity $\int_{\Omega} (s+i+r+d) d\Omega $ at all $t$ for both the uncoupled and coupled reconstructions and plot the results in Fig. \ref{fig:MassConsv}.

\par As we can see in the plot, the coupled DMD reconstruction is mass-conservative, for both the reconstruction and extrapolation periods, confirming the theoretical result. In contrast, the uncoupled DMD reconstruction fails to conserve mass, even failing during the training period. This provides a strong justification for the use of the coupled reconstruction formulation, as it is able to properly represent the underlying physics of the model.

\begin{figure}[ht!]
  \centering
 \includegraphics[width=.6\linewidth]{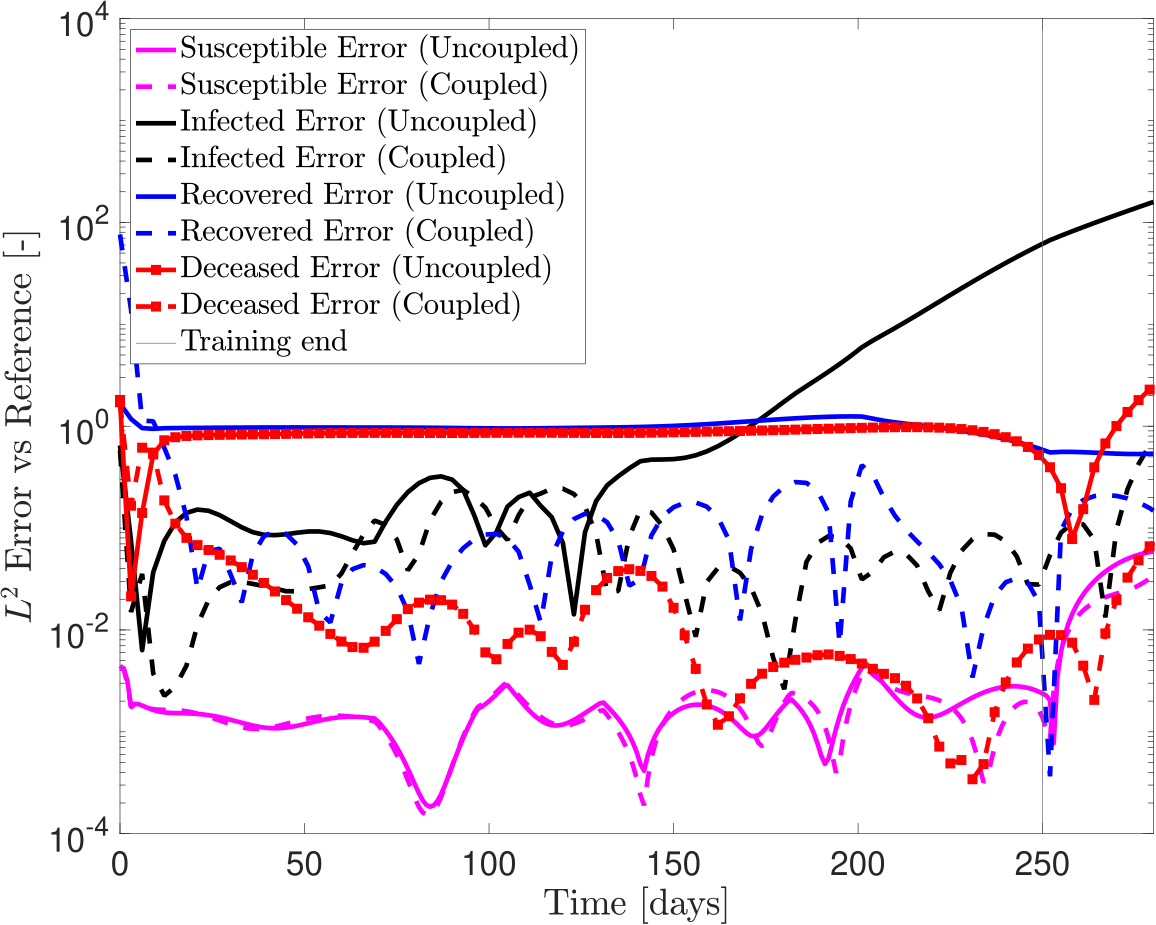}
  \caption{Error behavior in time for Georgia simulation. We see that the uncoupled DMD reconstructions fail for several compartments, giving unacceptable results. In contrast, the coupled DMD formulation is able to properly reconstruct the training dynamics, as well as provide reasonable extrapolations over a three-week interval, with the error compared to the reference simulation less than 1\% in several compartments as far as 15 days from the end of the training period.}
  \label{fig:GeorgiaError}
\end{figure}

\par
\begin{figure}[ht!]
  \centering
 \includegraphics[width=.6\linewidth]{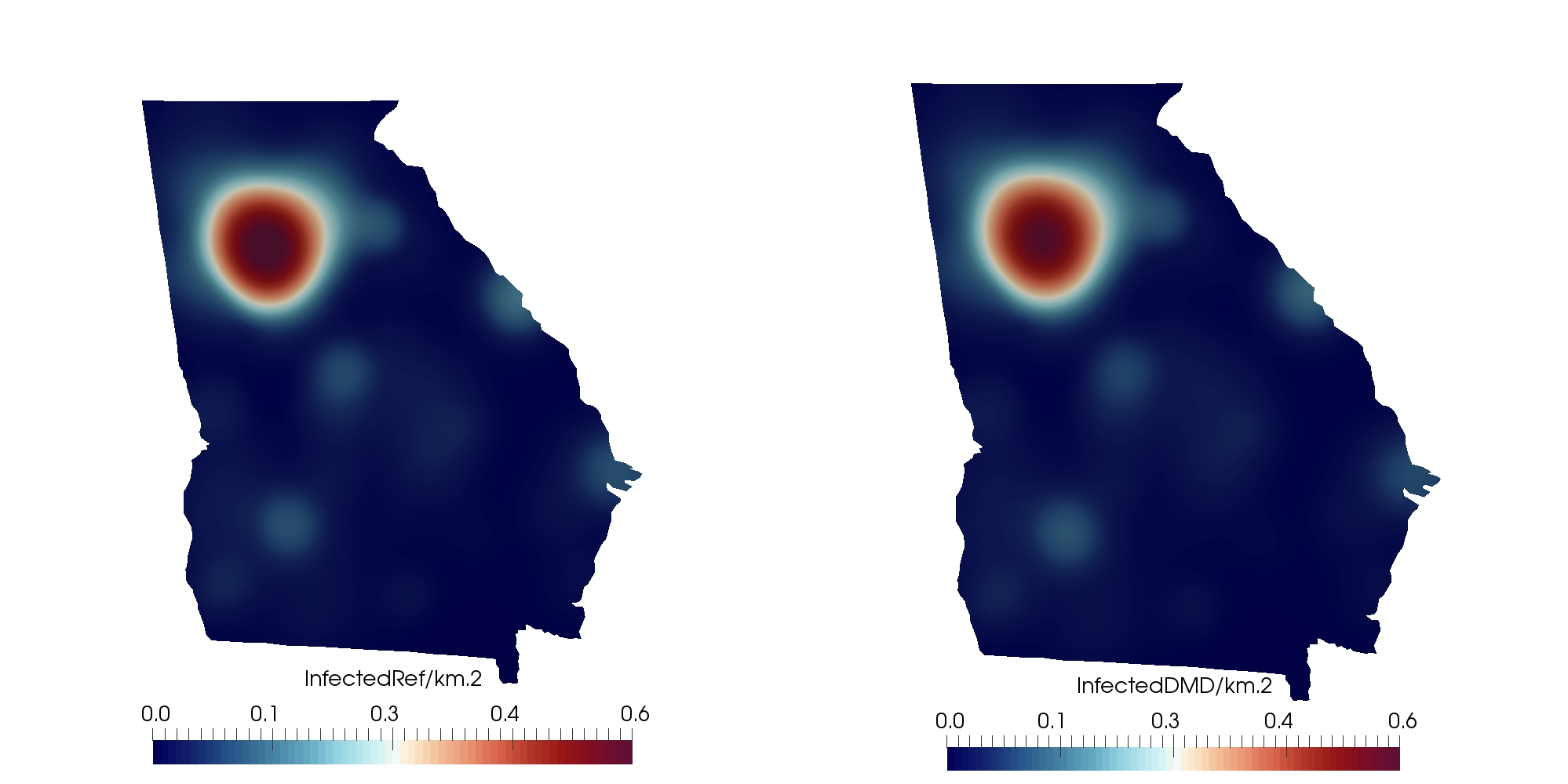}
  \caption{The infected compartment for Georgia, with the reference simulation (left) compared \textcolor{black}{with} the DMD reconstruction (right), 15 days after the end of the training period. We observe excellent agreement between the two quantities, with the relative $L^2$ error between them less than one percent.}
  \label{fig:GeorgiaDMDSideBySide}
\end{figure}

 We now turn our attention to the error behavior, defined as the relative $L^2$ error between the DMD reconstructions and the computed compartment for the reference simulations. The results are shown in Fig. \ref{fig:GeorgiaError}. As the figure shows, the uncoupled simulation is unable to provide meaningful results for the infected, deceased, and recovered compartments, performing only an acceptable reconstruction for the susceptible compartment. In contrast, the coupled reconstruction gives a good reconstruction of each compartment over the training period, with errors consistently lower than \textcolor{black}{10\%}, and also for a \textcolor{black}{significant} projection window. For the infected compartment, the error between reconstruction and reference is less than 1\% as far as 15 days from the end of the training period (Fig. \ref{fig:GeorgiaDMDSideBySide}). This is consistent with the theory, as the mass-conservation suggests that, as long as our reconstructions remain non-negative (as is the case here), the error in each compartment will remain under control.

\par Overall, we observe that the coupled DMD reconstruction provides reasonably accurate predictions over a short time interval, and is able to respect the underlying physics of the compartment. Indeed, even as the solution quality begins to deteriorate as we move further from the training period, the mass of the system remains conserved, suggesting that the overall system physics \textcolor{black}{is} well-incorporated into the DMD reconstruction. On the other hand, the compartment-by-compartment uncoupled DMD reconstruction neither respects the mass-conservation nor provides acceptable reconstructions and/or predictions. For this example, we see that the coupled DMD dynamics may indeed be applied for both signal reconstruction and short-term predictions, even when uncoupled DMD performs poorly.

%% file: sections/6_AM.tex
\subsection{Additive-manufacturing inspired nonlinear thermal and phase change problem}

In this section, we consider a two-dimensional problem inspired by additive manufacturing which \textcolor{black}{comprises} a layer of powder heated by a laser. As the powder reaches its melting point, it becomes a liquid, whereby upon cooling it becomes a solid. When reheated, the solid may again become liquid; however, the change from powder to liquid is irreversible. 
\par Mathematically, we consider the model used in \cite{viguerie2021numerical}, which consists of a partial differential equation for the temperature $T$ in a domain $\Omega$ and three ordinary differential equations at each point in $\Omega$ for the quantities $\phi_p$, $\phi_l$, and $\phi_s$: the phase fractions of the powder, liquid, and solid quantities respectively. The equations then read:

\begin{alignat}{2}
\rho c_p \frac{\partial T}{\partial t} - \nabla \cdot \left( \kappa \nabla T\right) &= L \qquad &\text{ in } \Omega \times \lbrack 0, T_{end} \rbrack,\label{temperature} \\
\frac{d \phi_p }{d t} &=
 -\frac{S_p}{T_f-T_s} \frac{\partial T}{\partial t} \phi_p \qquad &\text{ at each $x \in $  } \Omega \times \lbrack 0, T_{end}, \rbrack,\label{powder} \\
 \frac{d \phi_l }{d t} &=
 \frac{S_p}{T_f-T_s} \frac{\partial T}{\partial t} \phi_p  + \frac{S_s}{T_f-T_s} \frac{\partial T}{\partial t} \phi_s - \frac{S_l}{T_s-T_f} \frac{\partial T}{\partial t} \phi_l   \qquad &\text{ at each $x \in $  } \Omega \times \lbrack 0, T_{end}, \rbrack, \label{liquid}\\
  \frac{d \phi_s }{d t} &=
-\frac{S_s}{T_f-T_s} \frac{\partial T}{\partial t} \phi_s + \frac{S_l}{T_s-T_f} \frac{\partial T}{\partial t} \phi_l   \qquad &\text{ at each $x \in $  } \Omega \times \lbrack 0, T_{end}, \label{solid} \rbrack,
 \end{alignat}
\textcolor{black}{complemented by suitable} initial and boundary conditions. $T_s$ is the temperature where melting begins, and $T_f$ the temperature where melting is complete. $\rho$ is the material density, $c_p$ the specific heat, and $\kappa$ the thermal conductivity. These parameters depend on both the temperature field and material phase. This dependence is understood if not explicitly denoted. For these values, we base our simulation on stainless steel 316L, a common alloy used in additive manufacturing. More information about the material parameters and their specific relationship to temperature and material phase may be found in \cite{viguerie2021numerical, Mills2002}. The function $L$ is the laser function, defined in this work as a heat source, as done in  \cite{kollmannsberger2018hierarchical, kollmannsberger2019accurate}.
\par The functions $S_p$, $S_l$, and $S_s$ are sigmoid functions selected to satisfy the following properties:
\begin{align}
    S_p &=\begin{cases} 1 \text{ if  $\frac{\partial T}{\partial t} >0,\, T> T_s$ } \\ 0 \text{ else.}\end{cases}\\
S_l &=\begin{cases} 1 \text{ if  $\frac{\partial T}{\partial t} <0,\, T< T_f$  } \\ 0 \text{ else.}\end{cases}\\
S_s &=\begin{cases} 1 \text{ if  $\frac{\partial T}{\partial t} >0,\, T> T_s$  } \\ 0 \text{ else.}\end{cases}
\end{align}

Letting $\mathbf{u} = \lbrack T,\,\phi_p,\,\phi_l,\,\phi_s\rbrack^T $ and $\textcolor{black}{\boldsymbol{\zeta}}=\lbrack \kappa,\,S_p,\,S_l,\,S_s \rbrack^T$, if

\begin{equation}
   \mathbf{A} = \begin{bmatrix}
c_p \rho & 0 & 0  &0 \\
 0 & I & 0  & 0 \\
0 & 0 & I & 0 \\
0 & 0 & 0 & I \\
\end{bmatrix}
\end{equation}
\begin{equation}
   \mathbf{B} = \begin{bmatrix}
0 & 0 & 0 & 0 \\
0 & \frac{S_p}{T_f-T_s} \frac{\partial T}{\partial t} & 0 & 0 \\
0 & -\frac{S_p}{T_f-T_s} \frac{\partial T}{\partial t} & \frac{S_l}{T_s-T_f} \frac{\partial T}{\partial t} &- \frac{S_s}{T_s-T_f} \frac{\partial T}{\partial t} \\
0 & 0 & -\frac{S_l}{T_s-T_f} \frac{\partial T}{\partial t} &  \frac{S_s}{T_s-T_f} \frac{\partial T}{\partial t}  \\
\end{bmatrix}
\end{equation}

\begin{equation}
   \boldsymbol{\kappa} = \begin{bmatrix}
\boldsymbol\kappa_k & 0 & 0 & 0  \\
0  & 0 & 0 & 0\\
0  & 0 & 0 & 0\\
0  & 0 & 0 & 0\\
\end{bmatrix}
\end{equation}
\begin{equation}
   \boldsymbol{\kappa_k} = \begin{bmatrix}
 \kappa_{xx} & \kappa_{xy}\\
 \kappa_{yx} & \kappa_{yy} 
\end{bmatrix}
\end{equation}
\begin{equation}
\mathcal{N}(\mathbf{u},\textcolor{black}{\boldsymbol{\zeta}}) =  \mathbf{B} \mathbf{u} - \nabla \cdot \left( \boldsymbol{\kappa} \nabla \mathbf{u}\right)
\end{equation}

\begin{equation}
   \mathbf{f} = \begin{bmatrix}
L\\
0 \\
0 \\
0 \\
\end{bmatrix},
\end{equation}
we may write the system in a slightly altered version of the form given by \eqref{eq:strongVector}:
\begin{equation}\begin{split}
    \mathbf{A}\dfrac{\partial \mathbf{u}}{\partial t} + \mathcal{N}(\mathbf{u};\textcolor{black}{\boldsymbol{\zeta}}) =\mathbf{f}, \text{ \hspace{2cm} in $\Omega \times (0, T]$}. \\  
    \label{eq:modifiedStrongVector}
\end{split}\end{equation}

By summing \eqref{powder}-\eqref{solid}, one observes that:
$$ \frac{d }{dt }\lbrack \phi_p + \phi_l + \phi_s \rbrack=0 $$
at each $(\mathbf{x},t)$. By initializing $\phi_p=1$, $\phi_l=0$, $\phi_s=0$ (consistent with the physics), then expect that, at all $(\mathbf{x},t)$: 
$$ \phi_p + \phi_l + \phi_s=1.$$
It follows then that the system satisfies the hypotheses of Theorem 2; the phase fractions satisfy a mass conservation condition, while the temperature equation \eqref{temperature}, coupled to the phase-fraction dynamics, is separate from the compartmental system.

\subsubsection{Problem Setup and Simulation}
We consider a square domain assumed to be 400 microns in height and width, over a time interval of 7.5 milliseconds. We activate \textcolor{black}{the} laser function, defined as:
\begin{align}\label{laser}
    L(x,y) &= 9\cdot 10^6 \exp\left( \frac{ - (x-x_c)^4 - (y-y_c)^4 }{10^{-8}} \right),
\end{align}
where $(x_c,y_c)$ are 200 and 400 microns, respectively. The laser is activated for the first 1.5 milliseconds of the time period, and then switched off for the remainder, for the cooling phase.
\par In space, we discretize with $\mathbb{P}^1$ finite elements for the temperature and phase fields, over a mesh with size $h=8$ microns. For the boundary conditions, we fix the temperature along the bottom of the domain as 293.15 K, and in time, we use an implicit BE scheme with $\Delta t=0.0125$ milliseconds. These spatial and temporal discretizations were selected after mesh convergence analyses showed that an adequate level of spatial and temporal convergence \textcolor{black}{were} reached. \begin{figure}[ht!]
  \centering
 \includegraphics[width=.75\linewidth]{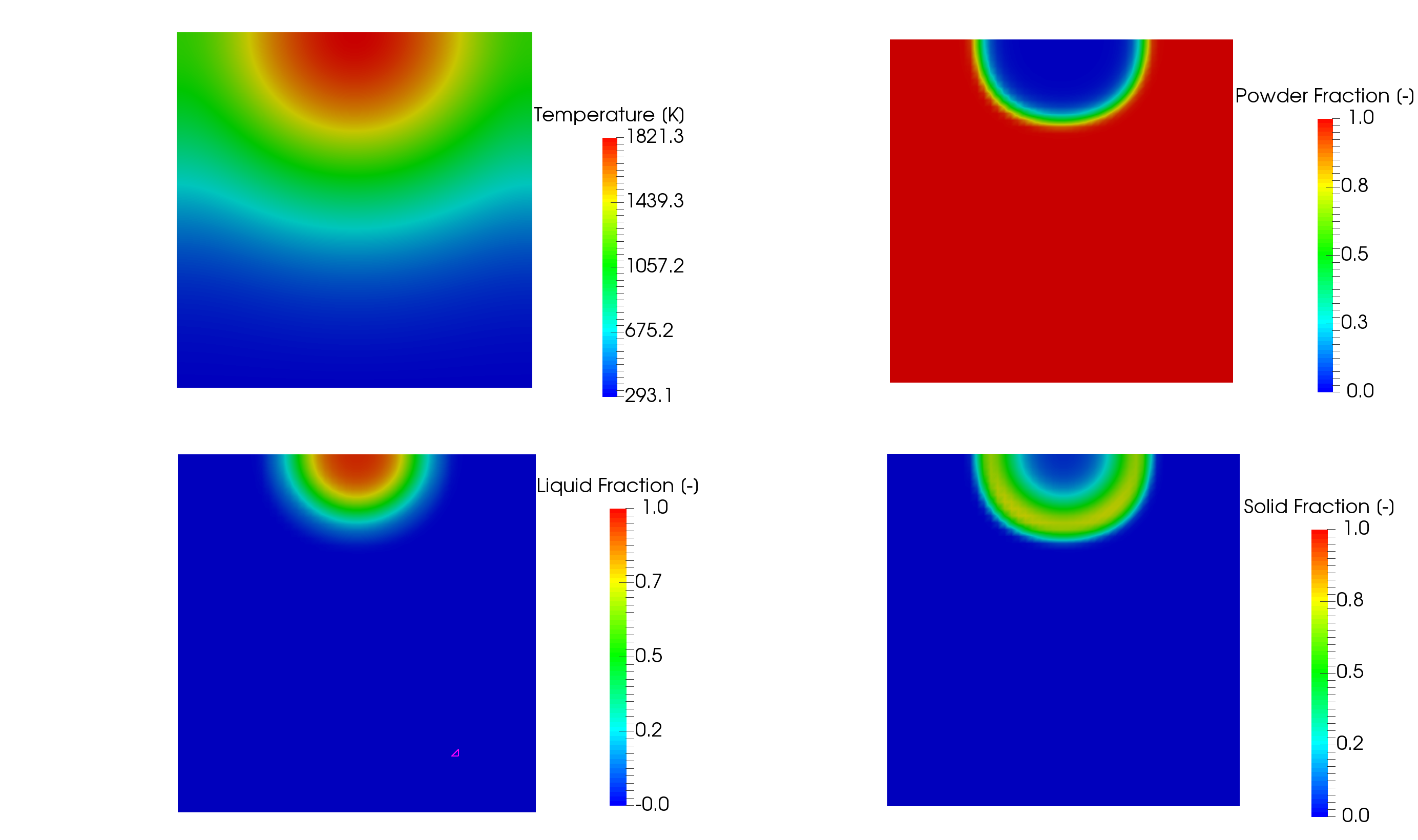}
  \caption{A snapshot at $t=1.81$ milliseconds for the additive manufacturing problem. Top-left: the temperature field, defined by a laser heating at the top of the domain. This laser, then heats the powder fraction (top right), which melts into a liquid (bottom left). Upon cooling, the liquid becomes a solid (bottom right).}
  \label{fig:AdditiveProblemDepiction}
\end{figure}
\par
\begin{figure}[ht!]
  \centering
 \includegraphics[width=\linewidth]{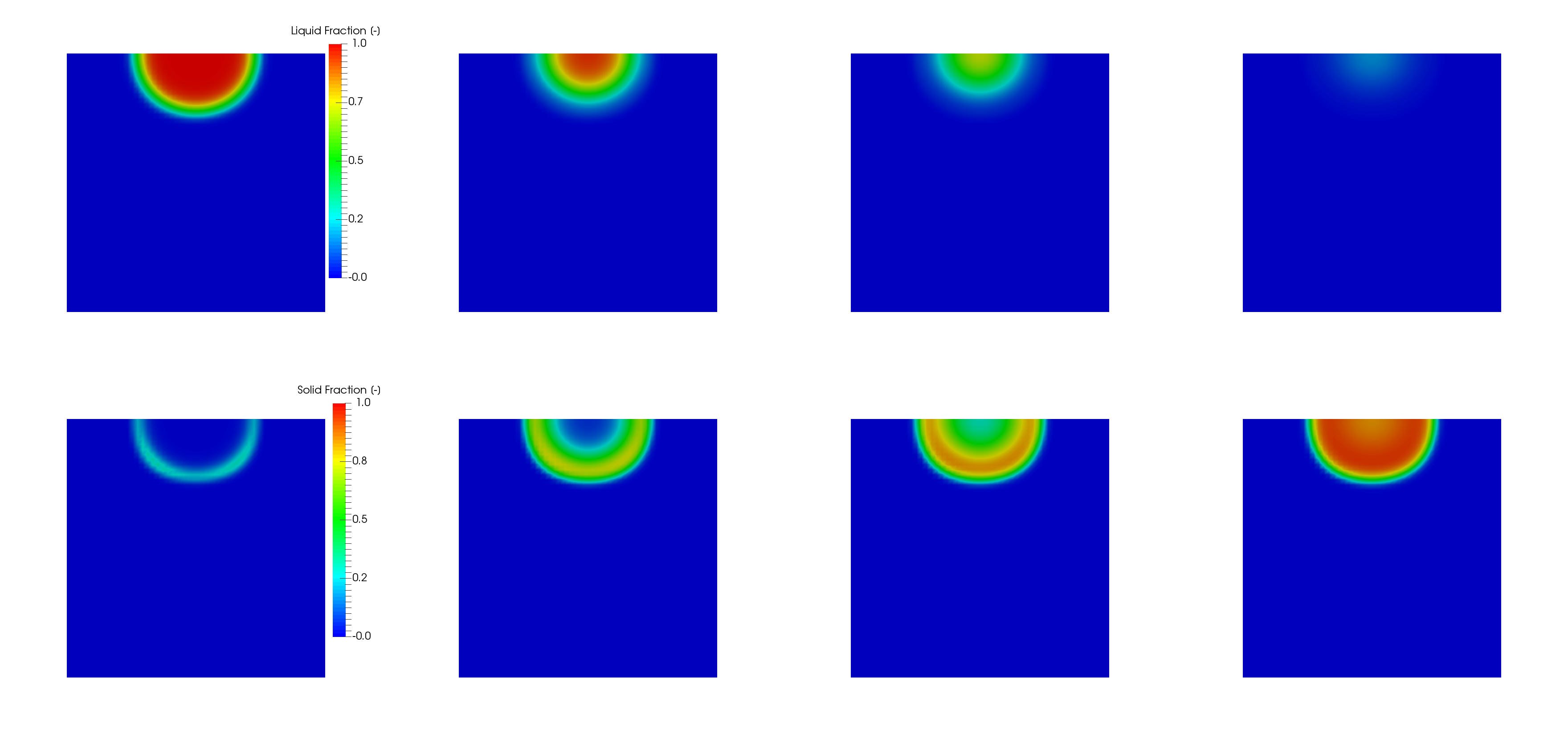}
  \caption{The phase change process from liquid to solid in more detail, shown at $t=1.56,\,1.81,\,1.94,\,2.19$ milliseconds. On the top row, we show the liquid fraction at four instances in time, with the bottom row depicting the solid fraction at the same four instances. As the material cools, the liquid begins to solidify, beginning first with the outer areas, before cooling and reaching a near-complete solid in the fourth instance shown. }
  \label{fig:SolidLiquid}
\end{figure}

 In Fig. \ref{fig:AdditiveProblemDepiction}, we show the solutions of the system \eqref{temperature}-\eqref{solid} at an instance in time. We see that the laser, defined at the top of the domain, provides a concentrated heat source, which melts the powder phase. At the instance shown, the laser has been deactivated, and we see the solidification process in action; while some areas of the melted powder remain hot enough such that the liquid phase dominates, the outer area has cooled enough to solidify. In Fig. \ref{fig:SolidLiquid}, we show this cooling/solidification process in more detail, showing the liquid (top) and solid (bottom) phases at four different instances in time, observing the change from liquid into solid. 

\subsubsection{Additive Manufacturing Problem, DMD Results}
We perform a similar analysis as in the previous example. We first compute the reference solution over the full time interval. Then, we define a training limit and reconstruct the DMD solutions for both the coupled and uncoupled DMD methods. In this instance, we defined the training limit to be $t=3.75$ seconds, halfway through the simulation. As before, we are interested in seeing how the coupled and uncoupled DMD compare, in terms of both their error behavior and their ability to properly respect the physics (e.g., mass conservation).
\begin{figure}[ht!]
  \centering
 \includegraphics[width=.75\linewidth]{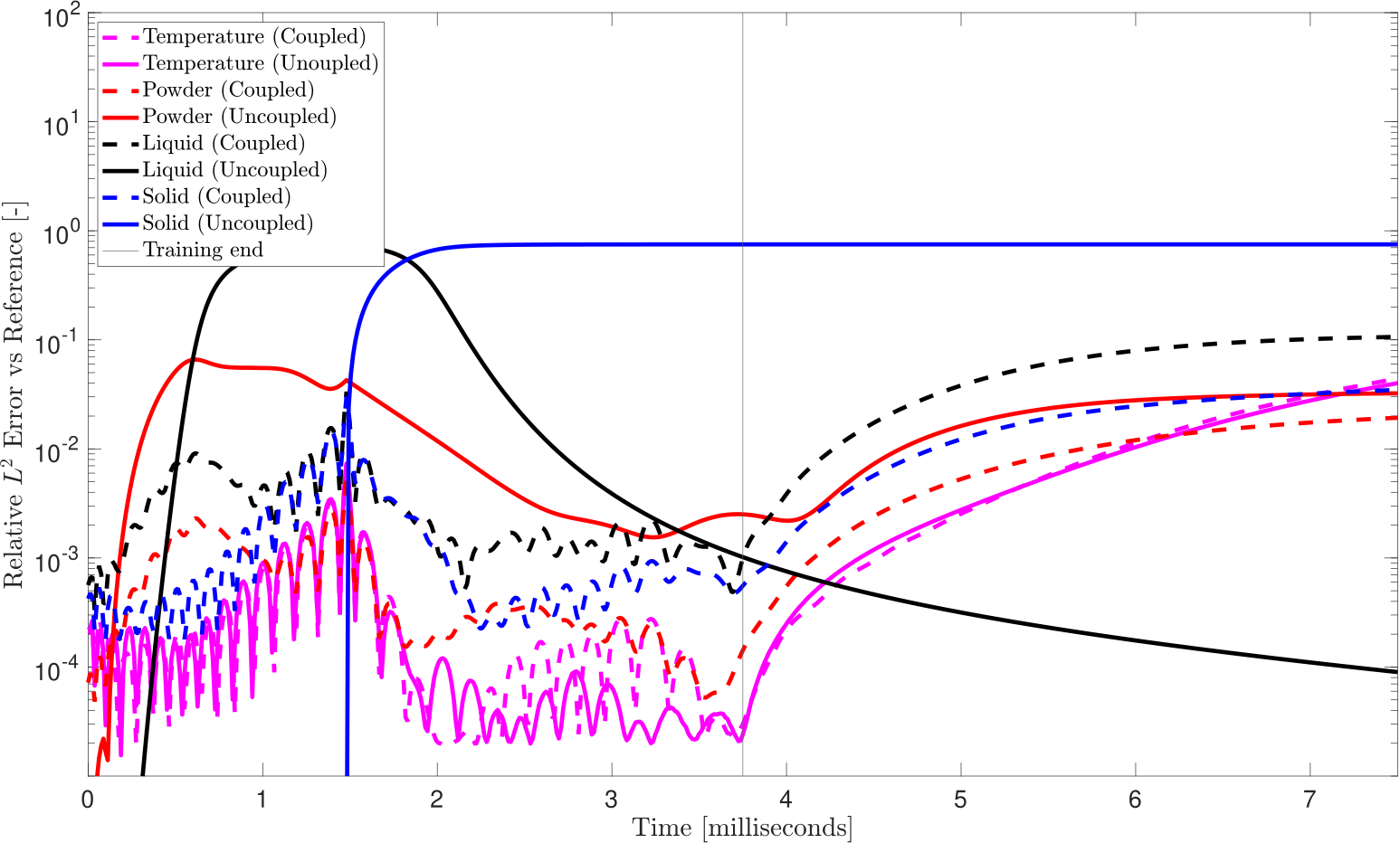}
  \caption{Error behavior in time for the additive manufacturing simulation. We see very similar error performance for both the coupled and uncoupled DMD for the temperature field, suggesting a relative independence from the phase compartment fields. However, for the phase fractions, the uncoupled DMD completely fails, with only the powder compartment showing reasonable behavior. The liquid and solid compartments, perhaps due to being initialized at zero, fail for both reconstruction and prediction. However, when applying the coupled DMD reconstruction, this effect is not an issue. }
  \label{fig:Additive}
\end{figure}
\par In Fig. \ref{fig:Additive} we plot the error behavior in time for the temperature and phase fraction fields for both the coupled and uncoupled cases. For the temperature field, the error behavior for both reconstructions is very similar. This is unsurprising, as the temperature field is somewhat independent of the compartmental physics, and therefore can be reconstructed with reasonable accuracy without considering the phase fractions. However, it is worth noting that, when considered jointly with the phase fractions, the error behavior shows little difference. We note that, in both the coupled and uncoupled case, the temperature error performance is very strong; it remains under 1\% for a long period of time after the end of the training period, and is still under 10\% at the end of the time interval. 

\par When considering the error in the phase fraction compartments, we see a clear advantage when using the coupled approach. The error remains low for both the reconstruction and prediction period for each phase fraction for coupled DMD. We observe errors under 1\% for the training period, and, as with the temperature field, they remain low throughout the time interval, even long after the training period is finished. However, for the uncoupled case, we observe failure in both the reconstruction and prediction. The solid and liquid phases, perhaps due to their initial conditions, remain near zero for the entirety of the time interval. In this sense, the long-term error behavior for the liquid compartment, while superficially appearing acceptable, is misleading; as the physical compartment tends to zero, the error appears acceptable. However, during the period in which the liquid compartment displays nonzero behavior, the uncoupled DMD fails to reproduce \textcolor{black}{this} dynamics, and the resulting error during this time is large. 

\begin{figure}[ht!]
  \centering
 \includegraphics[width=.6\linewidth]{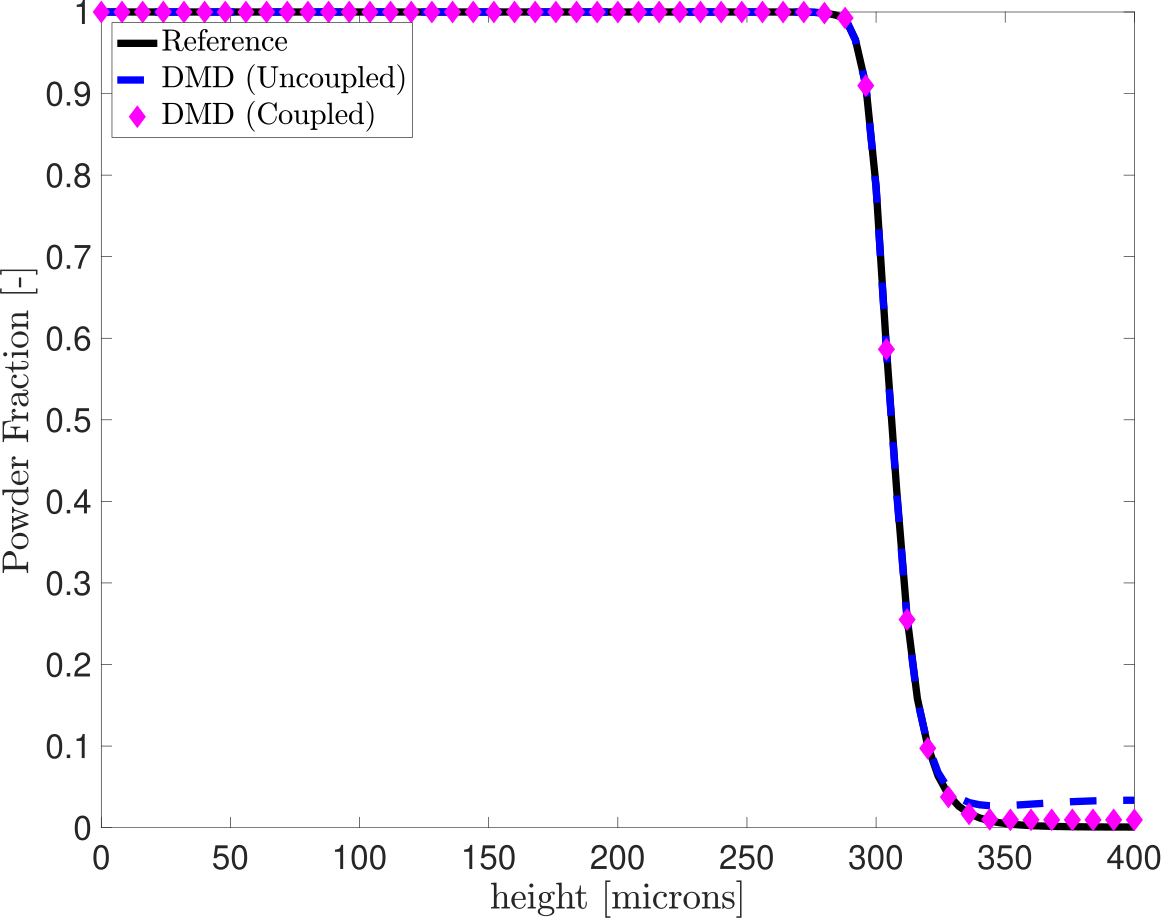}
  \caption{Melt pool profile for the reference case and the two different DMD approaches. We see reasonable agreement with the reference for both DMD methods; however, the coupled DMD approach nonetheless gives a demonstrably better result, matching the reference nearly perfectly, while the uncoupled DMD solution deviates more from the reference.  }
  \label{fig:MeltPool}
\end{figure}
\par We next examine the melt pool depth, often a quantity of much interest in simulations of additive manufacturing \cite{viguerie2021numerical}. We measure this by observing the powder phase fraction along the vertical centerline for the reference, coupled, and uncoupled DMD solutions. We plot these results in Fig. \ref{fig:MeltPool}. In this case, both the coupled and uncoupled DMD provide reasonable solutions (as expected given the error behavior of the powder fraction in Fig. \ref{fig:Additive}). However, the coupled approach nonetheless demonstrates superior performance, as the coupled approach shows noticeably more deviation from the reference solution towards the top of the domain.

\begin{figure}[ht!]
  \centering
 \includegraphics[width=.6\linewidth]{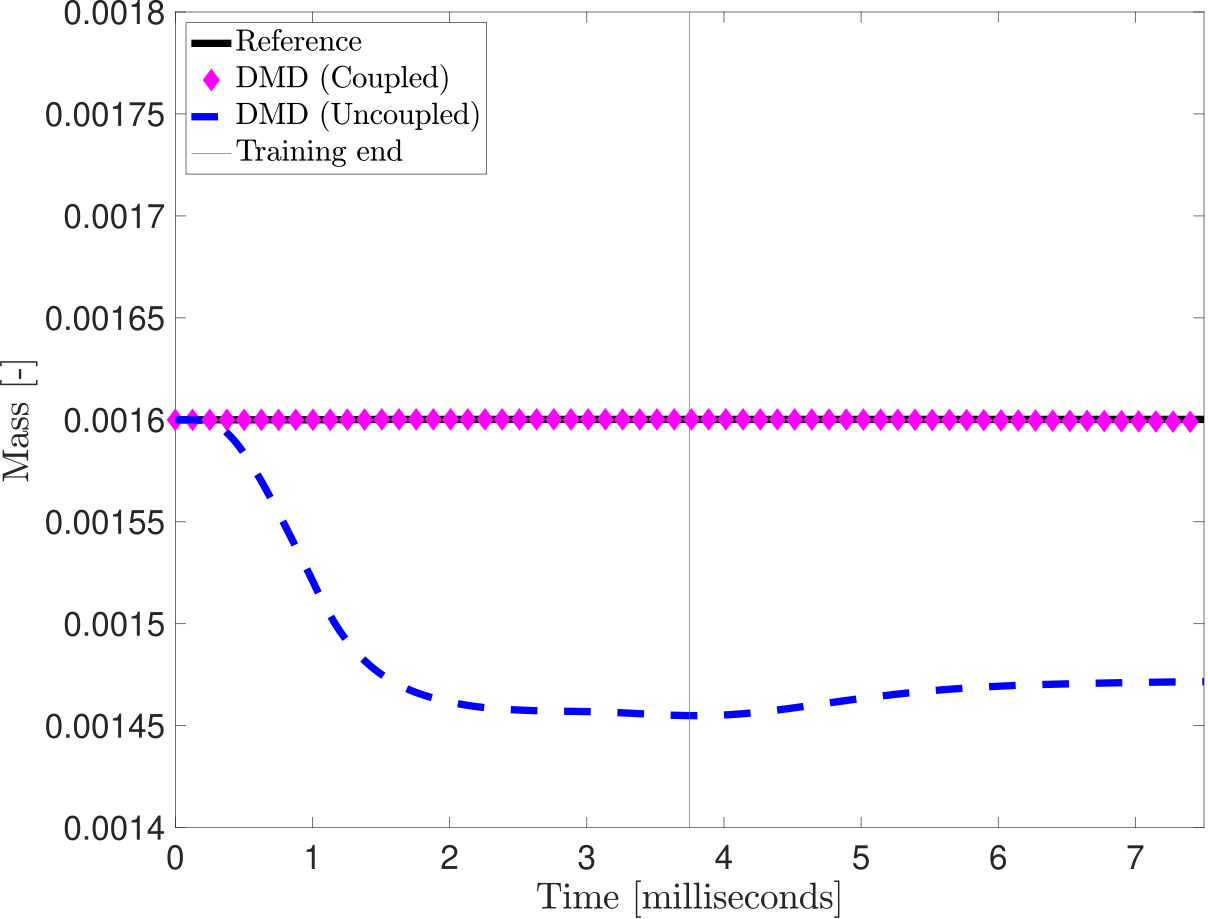}
  \caption{Mass conservation for the additive manufacturing problem. The mass is properly conserved with the coupled DMD analysis, confirming our theory. However, as in the SIRD model, the uncoupled DMD does not conserve mass, and the mass loss is quite pronounced in both the reconstruction and prediction periods. }
  \label{fig:AdditiveMass}
\end{figure}

\par In Fig. \ref{fig:AdditiveMass}, we \textcolor{black}{finally} examine the mass-conservation between the reference computation and the different DMD methods. As in the previous example, the coupled DMD reconstruction is mass-conservative throughout the considered interval, consistent with the theory. We note additionally that, as this model also incorporates a temperature field, independent of the compartment structure, this example is a numerical confirmation of Theorem 2. We again see that, when using an uncoupled DMD approach, the mass is not conserved.

%% file: sections/8_conclusions.tex
\section{Conclusions}
\label{sec:conclusions}
\par 
In this work, we examined two approaches for the dynamic mode decomposition method on compartmental PDE systems: the uncoupled approach, in which DMD is applied to each compartment separately, and the coupled approach, in which DMD is applied to all compartments together. The need for this study was motivated by the inconsistent performance observed for DMD as a predictive tool \textcolor{black}{in a number of situations}. We established theoretically that if a compartmental system is mass-conservative, as is often the case, we may expect DMD performed on the coupled system to respect this property, even when extrapolated in time. It was also shown that if this coupling holds only for a subset of the considered variables, we still nonetheless expect this property to hold when DMD is applied to the entire system. Based on these results, the error behavior may be improved as long as the compartments of the reconstructed system remain non-negative. 
\par We proceeded to show, through numerical examples, that coupled DMD can provide accurate reconstructions, even in cases when uncoupled DMD fails to give useful results, thus confirming our theoretical results. For an epidemiological model of COVID-19, it was shown that DMD provides accurate short-term predictions up to several weeks after the end of the training window. Further, the theoretical result concerning mass conservation was confirmed, and the error behavior was significantly better when compared to the uncoupled case, as expected. We proceeded to look at another application, motivated by additive manufacturing, which considered the evolution of a temperature field and the resulting change of phase between the powder, liquid, and solid states. This further confirmed our theory, and the uncoupled DMD was able to show good reconstructive and predictive ability and maintain the important mass-conservation properties.
\par The dynamic mode decomposition and applied Koopman theory is a very active area of modern research, with important works being published at an incredible rate. Such works include both theoretical analysis of DMD, as well as practical new algorithms and applications. Improving on the work shown here can be extended in both directions. In the application and practical setting, such as high order DMD \cite{le2017higher}, parameterized DMD, bagging for DMD \cite{sashidhar2021bagging}, multilevel DMD \cite{dylewsky2020dynamic}, adaptive DMD \cite{GVBRC2021}, and examining the performance of the coupled-vs.-uncoupled formulations is an important and interesting future path. Fast algorithms which may help alleviating possible additional costs of coupled DMD are also an important area of study; for systems with a very large number of compartments, the cost and memory requirements of coupled DMD may become prohibitive. Systematic ways to quantify the coupling such that one may reduce the cost, by, for instance, determining if one may obtain the advantages of coupled DMD by performing it only on a subset of compartments, are desired.  On the theoretical side, there are several mathematical results that we believe can be proven rigorously. As shown here, if one can assume non-negativity of the compartments in the DMD reconstruction, then the error of these reconstructions may remain bounded. However, in general, the reconstruction may not be non-negative, and as such, a formulation of DMD  \textcolor{black}{that} could guarantee this property would potentially prove a valuable tool in using DMD for longer-term time extrapolations.  

\section*{Acknowledgements}
This research was financed in part by the Coordena\c{c}\~ao de Aperfei\c{c}oamento de Pessoal de N\'ivel Superior - Brasil (CAPES) - Finance Code 001. This research has also received funding from CNPq and FAPERJ. A. Reali was partially supported  by the Italian Ministry of University and Research (MIUR) through the PRIN project XFAST-SIMS (No. 20173C478N).

\section*{Conflict of interest}
The authors declare that they have no conflict of interest.